\documentclass[preptrint, 12pt]{elsarticle}

    \usepackage{graphicx}
    \usepackage[table,xcdraw]{xcolor}
    \usepackage{tikz}
    \usepackage{makecell}
    \usepackage{dsfont}

    \usepackage{hyperref}

    \usepackage{amsmath}

    \usepackage{graphicx}

    \usepackage{svg}

    \usepackage{caption}

    \usepackage[Export]{adjustbox} 
    \adjustboxset{max size={0.9\linewidth}{0.9\paperheight}}
    \usepackage{float}
    \usepackage{xcolor} 
    \usepackage{adjustbox}
    \usepackage{enumerate} 
    \usepackage{geometry} 
    \usepackage{amsmath} 
    \usepackage{amssymb} 
    \usepackage{textcomp} 
    \AtBeginDocument{%
        \def\PYZsq{\textquotesingle}
    }
    \usepackage{upquote} 
    \usepackage{eurosym} 
    \usepackage[mathletters]{ucs} 
    \usepackage{fancyvrb} 
    \usepackage{bera}
    \usepackage{grffile} 
    \makeatletter 
    \def\Gread@@xetex#1{%
      \IfFileExists{"\Gin@base".bb}%
      {\Gread@eps{\Gin@base.bb}}%
      {\Gread@@xetex@aux#1}%
    }
    \makeatother

    \usepackage{hyperref}
    \usepackage{longtable} 
    \usepackage{booktabs}  
    \usepackage[inline]{enumitem} 
    \usepackage[normalem]{ulem} 
    \usepackage{mathrsfs}
    
    \usepackage[perpage]{footmisc} 

    \definecolor{urlcolor}{rgb}{0,.145,.698}
    \definecolor{linkcolor}{rgb}{.71,0.21,0.01}
    \definecolor{citecolor}{rgb}{.12,.54,.11}

    \definecolor{ansi-black}{HTML}{3E424D}
    \definecolor{ansi-black-intense}{HTML}{282C36}
    \definecolor{ansi-red}{HTML}{E75C58}
    \definecolor{ansi-red-intense}{HTML}{B22B31}
    \definecolor{ansi-green}{HTML}{00A250}
    \definecolor{ansi-green-intense}{HTML}{007427}
    \definecolor{ansi-yellow}{HTML}{DDB62B}
    \definecolor{ansi-yellow-intense}{HTML}{B27D12}
    \definecolor{ansi-blue}{HTML}{208FFB}
    \definecolor{ansi-blue-intense}{HTML}{0065CA}
    \definecolor{ansi-magenta}{HTML}{D160C4}
    \definecolor{ansi-magenta-intense}{HTML}{A03196}
    \definecolor{ansi-cyan}{HTML}{60C6C8}
    \definecolor{ansi-cyan-intense}{HTML}{258F8F}
    \definecolor{ansi-white}{HTML}{C5C1B4}
    \definecolor{ansi-white-intense}{HTML}{A1A6B2}
    \definecolor{ansi-default-inverse-fg}{HTML}{FFFFFF}
    \definecolor{ansi-default-inverse-bg}{HTML}{000000}

    \DefineVerbatimEnvironment{Highlighting}{Verbatim}{commandchars=\\\{\}}

    \newcommand{\GreenCircle}[1][blue!40!green,fill=blue!40!green]{\tikz[baseline=-0.5ex]\draw[#1,radius=3pt] (0,0) circle ;}%
    \newcommand{\OrangeCircle}[1][red!30!yellow,fill=red!30!yellow]{\tikz[baseline=-0.5ex]\draw[#1,radius=3pt] (0,0) circle ;}%
    \newcommand{\RedCircle}[1][black!10!red,fill=black!10!red]{\tikz[baseline=-0.5ex]\draw[#1,radius=3pt] (0,0) circle ;}%
    \newcommand{\GrayCircle}[1][gray,fill=gray]{\tikz[baseline=-0.5ex]\draw[#1,radius=3pt] (0,0) circle ;}%
    
    \let\Oldtex\TeX
    \let\Oldlatex\LaTeX
    \renewcommand{\TeX}{\textrm{\Oldtex}}
    \renewcommand{\LaTeX}{\textrm{\Oldlatex}}

\newcommand{\sherpa}[1]{{\color{black}#1}}    
\newcommand{\ugr}[1]{{\color{black}#1}}  
\newcommand{\paco}[1]{{\color{black}#1}}
\newcommand{\eugenio}[1]{{\color{black}#1}}
\newcommand{\nuria}[1]{{\color{black}#1}}

\newcommand{\correcciones}[1]{{\color{black}#1}}

\newcommand{\Hquad}{\hspace{0.5em}}

\makeatletter
\def\PY@reset{\let\PY@it=\relax \let\PY@bf=\relax%
    \let\PY@ul=\relax \let\PY@tc=\relax%
    \let\PY@bc=\relax \let\PY@ff=\relax}
\def\PY@tok#1{\csname PY@tok@#1\endcsname}
\def\PY@toks#1+{\ifx\relax#1\empty\else%
    \PY@tok{#1}\expandafter\PY@toks\fi}
\def\PY@do#1{\PY@bc{\PY@tc{\PY@ul{%
    \PY@it{\PY@bf{\PY@ff{#1}}}}}}}
\def\PY#1#2{\PY@reset\PY@toks#1+\relax+\PY@do{#2}}

\expandafter\def\csname PY@tok@w\endcsname{\def\PY@tc##1{\textcolor[rgb]{0.73,0.73,0.73}{##1}}}
\expandafter\def\csname PY@tok@c\endcsname{\let\PY@it=\textit\def\PY@tc##1{\textcolor[rgb]{0.25,0.50,0.50}{##1}}}
\expandafter\def\csname PY@tok@cp\endcsname{\def\PY@tc##1{\textcolor[rgb]{0.74,0.48,0.00}{##1}}}
\expandafter\def\csname PY@tok@k\endcsname{\let\PY@bf=\textbf\def\PY@tc##1{\textcolor[rgb]{0.00,0.50,0.00}{##1}}}
\expandafter\def\csname PY@tok@kp\endcsname{\def\PY@tc##1{\textcolor[rgb]{0.00,0.50,0.00}{##1}}}
\expandafter\def\csname PY@tok@kt\endcsname{\def\PY@tc##1{\textcolor[rgb]{0.69,0.00,0.25}{##1}}}
\expandafter\def\csname PY@tok@o\endcsname{\def\PY@tc##1{\textcolor[rgb]{0.40,0.40,0.40}{##1}}}
\expandafter\def\csname PY@tok@ow\endcsname{\let\PY@bf=\textbf\def\PY@tc##1{\textcolor[rgb]{0.67,0.13,1.00}{##1}}}
\expandafter\def\csname PY@tok@nb\endcsname{\def\PY@tc##1{\textcolor[rgb]{0.00,0.50,0.00}{##1}}}
\expandafter\def\csname PY@tok@nf\endcsname{\def\PY@tc##1{\textcolor[rgb]{0.00,0.00,1.00}{##1}}}
\expandafter\def\csname PY@tok@nc\endcsname{\let\PY@bf=\textbf\def\PY@tc##1{\textcolor[rgb]{0.00,0.00,1.00}{##1}}}
\expandafter\def\csname PY@tok@nn\endcsname{\let\PY@bf=\textbf\def\PY@tc##1{\textcolor[rgb]{0.00,0.00,1.00}{##1}}}
\expandafter\def\csname PY@tok@ne\endcsname{\let\PY@bf=\textbf\def\PY@tc##1{\textcolor[rgb]{0.82,0.25,0.23}{##1}}}
\expandafter\def\csname PY@tok@nv\endcsname{\def\PY@tc##1{\textcolor[rgb]{0.10,0.09,0.49}{##1}}}
\expandafter\def\csname PY@tok@no\endcsname{\def\PY@tc##1{\textcolor[rgb]{0.53,0.00,0.00}{##1}}}
\expandafter\def\csname PY@tok@nl\endcsname{\def\PY@tc##1{\textcolor[rgb]{0.63,0.63,0.00}{##1}}}
\expandafter\def\csname PY@tok@ni\endcsname{\let\PY@bf=\textbf\def\PY@tc##1{\textcolor[rgb]{0.60,0.60,0.60}{##1}}}
\expandafter\def\csname PY@tok@na\endcsname{\def\PY@tc##1{\textcolor[rgb]{0.49,0.56,0.16}{##1}}}
\expandafter\def\csname PY@tok@nt\endcsname{\let\PY@bf=\textbf\def\PY@tc##1{\textcolor[rgb]{0.00,0.50,0.00}{##1}}}
\expandafter\def\csname PY@tok@nd\endcsname{\def\PY@tc##1{\textcolor[rgb]{0.67,0.13,1.00}{##1}}}
\expandafter\def\csname PY@tok@s\endcsname{\def\PY@tc##1{\textcolor[rgb]{0.73,0.13,0.13}{##1}}}
\expandafter\def\csname PY@tok@sd\endcsname{\let\PY@it=\textit\def\PY@tc##1{\textcolor[rgb]{0.73,0.13,0.13}{##1}}}
\expandafter\def\csname PY@tok@si\endcsname{\let\PY@bf=\textbf\def\PY@tc##1{\textcolor[rgb]{0.73,0.40,0.53}{##1}}}
\expandafter\def\csname PY@tok@se\endcsname{\let\PY@bf=\textbf\def\PY@tc##1{\textcolor[rgb]{0.73,0.40,0.13}{##1}}}
\expandafter\def\csname PY@tok@sr\endcsname{\def\PY@tc##1{\textcolor[rgb]{0.73,0.40,0.53}{##1}}}
\expandafter\def\csname PY@tok@ss\endcsname{\def\PY@tc##1{\textcolor[rgb]{0.10,0.09,0.49}{##1}}}
\expandafter\def\csname PY@tok@sx\endcsname{\def\PY@tc##1{\textcolor[rgb]{0.00,0.50,0.00}{##1}}}
\expandafter\def\csname PY@tok@m\endcsname{\def\PY@tc##1{\textcolor[rgb]{0.40,0.40,0.40}{##1}}}
\expandafter\def\csname PY@tok@gh\endcsname{\let\PY@bf=\textbf\def\PY@tc##1{\textcolor[rgb]{0.00,0.00,0.50}{##1}}}
\expandafter\def\csname PY@tok@gu\endcsname{\let\PY@bf=\textbf\def\PY@tc##1{\textcolor[rgb]{0.50,0.00,0.50}{##1}}}
\expandafter\def\csname PY@tok@gd\endcsname{\def\PY@tc##1{\textcolor[rgb]{0.63,0.00,0.00}{##1}}}
\expandafter\def\csname PY@tok@gi\endcsname{\def\PY@tc##1{\textcolor[rgb]{0.00,0.63,0.00}{##1}}}
\expandafter\def\csname PY@tok@gr\endcsname{\def\PY@tc##1{\textcolor[rgb]{1.00,0.00,0.00}{##1}}}
\expandafter\def\csname PY@tok@ge\endcsname{\let\PY@it=\textit}
\expandafter\def\csname PY@tok@gs\endcsname{\let\PY@bf=\textbf}
\expandafter\def\csname PY@tok@gp\endcsname{\let\PY@bf=\textbf\def\PY@tc##1{\textcolor[rgb]{0.00,0.00,0.50}{##1}}}
\expandafter\def\csname PY@tok@go\endcsname{\def\PY@tc##1{\textcolor[rgb]{0.53,0.53,0.53}{##1}}}
\expandafter\def\csname PY@tok@gt\endcsname{\def\PY@tc##1{\textcolor[rgb]{0.00,0.27,0.87}{##1}}}
\expandafter\def\csname PY@tok@err\endcsname{\def\PY@bc##1{\setlength{\fboxsep}{0pt}\fcolorbox[rgb]{1.00,0.00,0.00}{1,1,1}{\strut ##1}}}
\expandafter\def\csname PY@tok@kc\endcsname{\let\PY@bf=\textbf\def\PY@tc##1{\textcolor[rgb]{0.00,0.50,0.00}{##1}}}
\expandafter\def\csname PY@tok@kd\endcsname{\let\PY@bf=\textbf\def\PY@tc##1{\textcolor[rgb]{0.00,0.50,0.00}{##1}}}
\expandafter\def\csname PY@tok@kn\endcsname{\let\PY@bf=\textbf\def\PY@tc##1{\textcolor[rgb]{0.00,0.50,0.00}{##1}}}
\expandafter\def\csname PY@tok@kr\endcsname{\let\PY@bf=\textbf\def\PY@tc##1{\textcolor[rgb]{0.00,0.50,0.00}{##1}}}
\expandafter\def\csname PY@tok@bp\endcsname{\def\PY@tc##1{\textcolor[rgb]{0.00,0.50,0.00}{##1}}}
\expandafter\def\csname PY@tok@fm\endcsname{\def\PY@tc##1{\textcolor[rgb]{0.00,0.00,1.00}{##1}}}
\expandafter\def\csname PY@tok@vc\endcsname{\def\PY@tc##1{\textcolor[rgb]{0.10,0.09,0.49}{##1}}}
\expandafter\def\csname PY@tok@vg\endcsname{\def\PY@tc##1{\textcolor[rgb]{0.10,0.09,0.49}{##1}}}
\expandafter\def\csname PY@tok@vi\endcsname{\def\PY@tc##1{\textcolor[rgb]{0.10,0.09,0.49}{##1}}}
\expandafter\def\csname PY@tok@vm\endcsname{\def\PY@tc##1{\textcolor[rgb]{0.10,0.09,0.49}{##1}}}
\expandafter\def\csname PY@tok@sa\endcsname{\def\PY@tc##1{\textcolor[rgb]{0.73,0.13,0.13}{##1}}}
\expandafter\def\csname PY@tok@sb\endcsname{\def\PY@tc##1{\textcolor[rgb]{0.73,0.13,0.13}{##1}}}
\expandafter\def\csname PY@tok@sc\endcsname{\def\PY@tc##1{\textcolor[rgb]{0.73,0.13,0.13}{##1}}}
\expandafter\def\csname PY@tok@dl\endcsname{\def\PY@tc##1{\textcolor[rgb]{0.73,0.13,0.13}{##1}}}
\expandafter\def\csname PY@tok@s2\endcsname{\def\PY@tc##1{\textcolor[rgb]{0.73,0.13,0.13}{##1}}}
\expandafter\def\csname PY@tok@sh\endcsname{\def\PY@tc##1{\textcolor[rgb]{0.73,0.13,0.13}{##1}}}
\expandafter\def\csname PY@tok@s1\endcsname{\def\PY@tc##1{\textcolor[rgb]{0.73,0.13,0.13}{##1}}}
\expandafter\def\csname PY@tok@mb\endcsname{\def\PY@tc##1{\textcolor[rgb]{0.40,0.40,0.40}{##1}}}
\expandafter\def\csname PY@tok@mf\endcsname{\def\PY@tc##1{\textcolor[rgb]{0.40,0.40,0.40}{##1}}}
\expandafter\def\csname PY@tok@mh\endcsname{\def\PY@tc##1{\textcolor[rgb]{0.40,0.40,0.40}{##1}}}
\expandafter\def\csname PY@tok@mi\endcsname{\def\PY@tc##1{\textcolor[rgb]{0.40,0.40,0.40}{##1}}}
\expandafter\def\csname PY@tok@il\endcsname{\def\PY@tc##1{\textcolor[rgb]{0.40,0.40,0.40}{##1}}}
\expandafter\def\csname PY@tok@mo\endcsname{\def\PY@tc##1{\textcolor[rgb]{0.40,0.40,0.40}{##1}}}
\expandafter\def\csname PY@tok@ch\endcsname{\let\PY@it=\textit\def\PY@tc##1{\textcolor[rgb]{0.25,0.50,0.50}{##1}}}
\expandafter\def\csname PY@tok@cm\endcsname{\let\PY@it=\textit\def\PY@tc##1{\textcolor[rgb]{0.25,0.50,0.50}{##1}}}
\expandafter\def\csname PY@tok@cpf\endcsname{\let\PY@it=\textit\def\PY@tc##1{\textcolor[rgb]{0.25,0.50,0.50}{##1}}}
\expandafter\def\csname PY@tok@c1\endcsname{\let\PY@it=\textit\def\PY@tc##1{\textcolor[rgb]{0.25,0.50,0.50}{##1}}}
\expandafter\def\csname PY@tok@cs\endcsname{\let\PY@it=\textit\def\PY@tc##1{\textcolor[rgb]{0.25,0.50,0.50}{##1}}}

\def\PYZus{\char`\_}

\def\PYZsh{\char`\#}

\def\PYZhy{\char`\-}
\def\PYZsq{\char`\'}
\def\PYZdq{\char`\"}

\makeatother

    \makeatletter
        \newbox\Wrappedcontinuationbox 
        \newbox\Wrappedvisiblespacebox 
        \newcommand*\Wrappedvisiblespace {\textcolor{red}{\textvisiblespace}} 
        \newcommand*\Wrappedcontinuationsymbol {\textcolor{red}{\llap{\tiny$\m@th\hookrightarrow$}}} 
        \newcommand*\Wrappedcontinuationindent {3ex } 
        \newcommand*\Wrappedafterbreak {\kern\Wrappedcontinuationindent\copy\Wrappedcontinuationbox} 
        \newcommand*\Wrappedbreaksatspecials {%
            \def\PYGZus{\discretionary{\char`\_}{\Wrappedafterbreak}{\char`\_}}%
            \def\PYGZob{\discretionary{}{\Wrappedafterbreak\char`\{}{\char`\{}}%
            \def\PYGZcb{\discretionary{\char`\}}{\Wrappedafterbreak}{\char`\}}}%
            \def\PYGZca{\discretionary{\char`\^}{\Wrappedafterbreak}{\char`\^}}%
            \def\PYGZam{\discretionary{\char`\&}{\Wrappedafterbreak}{\char`\&}}%
            \def\PYGZlt{\discretionary{}{\Wrappedafterbreak\char`\<}{\char`\<}}%
            \def\PYGZgt{\discretionary{\char`\>}{\Wrappedafterbreak}{\char`\>}}%
            \def\PYGZsh{\discretionary{}{\Wrappedafterbreak\char`\#}{\char`\#}}%
            \def\PYGZpc{\discretionary{}{\Wrappedafterbreak\char`\%}{\char`\%}}%
            \def\PYGZdl{\discretionary{}{\Wrappedafterbreak\char`\$}{\char`\$}}%
            \def\PYGZhy{\discretionary{\char`\-}{\Wrappedafterbreak}{\char`\-}}%
            \def\PYGZsq{\discretionary{}{\Wrappedafterbreak\textquotesingle}{\textquotesingle}}%
            \def\PYGZdq{\discretionary{}{\Wrappedafterbreak\char`\"}{\char`\"}}%
            \def\PYGZti{\discretionary{\char`\~}{\Wrappedafterbreak}{\char`\~}}%
        } 
        \newcommand*\Wrappedbreaksatpunct {%
            \lccode`\~`\.\lowercase{\def~}{\discretionary{\hbox{\char`\.}}{\Wrappedafterbreak}{\hbox{\char`\.}}}%
            \lccode`\~`\,\lowercase{\def~}{\discretionary{\hbox{\char`\,}}{\Wrappedafterbreak}{\hbox{\char`\,}}}%
            \lccode`\~`\;\lowercase{\def~}{\discretionary{\hbox{\char`\;}}{\Wrappedafterbreak}{\hbox{\char`\;}}}%
            \lccode`\~`\:\lowercase{\def~}{\discretionary{\hbox{\char`\:}}{\Wrappedafterbreak}{\hbox{\char`\:}}}%
            \lccode`\~`\?\lowercase{\def~}{\discretionary{\hbox{\char`\?}}{\Wrappedafterbreak}{\hbox{\char`\?}}}%
            \lccode`\~`\!\lowercase{\def~}{\discretionary{\hbox{\char`\!}}{\Wrappedafterbreak}{\hbox{\char`\!}}}%
            \lccode`\~`\/\lowercase{\def~}{\discretionary{\hbox{\char`\/}}{\Wrappedafterbreak}{\hbox{\char`\/}}}%
            \catcode`\.\active
            \catcode`\,\active 
            \catcode`\;\active
            \catcode`\:\active
            \catcode`\?\active
            \catcode`\!\active
            \catcode`\/\active 
            \lccode`\~`\~ 	
        }
    \makeatother

    \let\OriginalVerbatim=\Verbatim
    \makeatletter
    \renewcommand{\Verbatim}[1][1]{%
        \sbox\Wrappedcontinuationbox {\Wrappedcontinuationsymbol}%
        \sbox\Wrappedvisiblespacebox {\FV@SetupFont\Wrappedvisiblespace}%
        \def\FancyVerbFormatLine ##1{\hsize\linewidth
            \vtop{\raggedright\hyphenpenalty\z@\exhyphenpenalty\z@
                \doublehyphendemerits\z@\finalhyphendemerits\z@
                \strut ##1\strut}%
        }%
        \def\FV@Space {%
            \nobreak\hskip\z@ plus\fontdimen3\font minus\fontdimen4\font
            \discretionary{\copy\Wrappedvisiblespacebox}{\Wrappedafterbreak}
            {\kern\fontdimen2\font}%
        }%
        
        \Wrappedbreaksatspecials
        \OriginalVerbatim[#1,codes*=\Wrappedbreaksatpunct]%
    }
    \makeatother

    \definecolor{incolor}{HTML}{303F9F}
    \definecolor{outcolor}{HTML}{D84315}
    \definecolor{cellborder}{HTML}{CFCFCF}
    \definecolor{cellbackground}{HTML}{F7F7F7}
    
    \makeatletter
    \newcommand{\boxspacing}{\kern\kvtcb@left@rule\kern\kvtcb@boxsep}
    \makeatother
    \newcommand{\prompt}[4]{
        \ttfamily\llap{{\color{#2}[#3]:\hspace{3pt}#4}}\vspace{-\baselineskip}
    }

    \hypersetup{
      breaklinks=true,  
      colorlinks=true,
      urlcolor=urlcolor,
      linkcolor=linkcolor,
      citecolor=citecolor,
      }
    
\usepackage{amssymb}

\usepackage{threeparttable}

\usepackage{tabularx}

\usepackage{color}

\usepackage{graphicx}

\usepackage{subfig}

\graphicspath{{images/}}

\usepackage{url}

\usepackage{pgfplots}

\usepackage{caption}

\usepackage{color}

\usepackage{amsmath}

\usepackage{lscape}

\usepackage{todonotes}

\usepackage{comment}

\usepackage{soul}

\usepackage[inline]{enumitem}

\usepackage[breakable]{tcolorbox}
\usepackage{parskip} 

\usepackage{iftex}
\ifPDFTeX
	\usepackage[T1]{fontenc}
	\usepackage{mathpazo}
\else
	\usepackage{fontspec}
\fi

\journal{Information Fusion}
\let\today\relax
\makeatletter
\def\ps@pprintTitle{%
    \let\@oddhead\@empty
    \let\@evenhead\@empty
    \def\@oddfoot{\footnotesize\itshape
         {} \hfill\today}%
    \let\@evenfoot\@oddfoot
    }
\makeatother

\begin{document}

\begin{frontmatter}

\title{\ugr{Federated Learning and Differential Privacy: 
Software tools \correcciones{analysis}, the \sherpa{Sherpa.ai FL framework} and methodological guidelines for preserving data privacy}}


\author[1]{Nuria Rodr\'{i}guez-Barroso} \ead{rbnuria@ugr.es}
\author[2]{Goran Stipcich} \ead{g.stipcich@sherpa.ai}
\author[1]{Daniel Jim\'{e}nez-L\'{o}pez} \ead{dajilo@ugr.es}
\author[1]{Jos\'{e} Antonio Ruiz-Mill\'{a}n} \ead{jantonioruiz@ugr.es}
\author[1]{Eugenio Mart\'{i}nez-C\'{a}mara\corref{cor1}} \ead{emcamara@decsai.ugr.es}
\author[2]{Gerardo Gonz\'{a}lez-Seco} \ead{g.gonzalez@sherpa.ai}
\author[1]{M. Victoria Luz\'{o}n}
\ead{luzon@ugr.es}
\author[2]{Miguel \'{A}ngel Veganzones} \ead{ma.veganzones@sherpa.ai}
\author[1]{Francisco Herrera} \ead{herrera@decsai.ugr.es}

\cortext[cor1]{Corresponding author}
\address[1]{Andalusian Research Institute in Data Science and Computational Intelligence, University of Granada, Spain}
\address[2]{Sherpa.ai, Bilbao, Spain}

\begin{abstract}






\correcciones{The high demand of artificial intelligence services at the edges that also preserve data privacy has pushed the research on novel machine learning paradigms that fit these requirements. Federated learning has the ambition to protect data privacy through distributed learning methods that keep the data in its storage silos. Likewise, differential privacy attains to improve the protection of data privacy by measuring the privacy loss in the communication among the elements of federated learning. The prospective matching of federated learning and differential privacy to the challenges of data privacy protection has caused the release of several software tools that support their functionalities, but they lack a unified vision of these techniques, and a methodological workflow that supports their usage. Hence, we present \sherpa{the \texttt{Sherpa.ai} Federated Learning framework} that is built upon a holistic view of federated learning and differential privacy. It results from both the study of how to adapt the machine learning paradigm to federated learning, and the definition of  methodological guidelines for developing artificial intelligence services based on federated learning and differential privacy. We show how to follow the methodological guidelines with \sherpa{the \texttt{Sherpa.ai} Federated Learning framework} by means of a classification and a regression use cases.}

\end{abstract}

\begin{keyword}
federated learning \sep differential privacy \sep software framework \sep \sherpa{\texttt{Sherpa.ai} Federated Learning framework}
\end{keyword}

\end{frontmatter}

\section{Introduction}
\label{sec:introduction}

\correcciones{The last advances in fundamental and applied research in artificial intelligence (AI) has aroused  interest in industry and end users. This interest goes beyond the traditional centralised setting of AI, and nowadays there is a high demand of AI services at the edges.

One of the main pillars of AI is data, whose larger availability has boosted the progress of AI in the last years. However, data is a sensitive element, especially when it describes users' personal features, such  as clinical or financial data. This sensitive nature of personal data has raised the awareness of end users on data privacy protection, promoting the publication of legal frames \cite{eu_prot_datos} and recommendations for developing AI services that preserve data privacy  \cite{eu_ia_fiable}.

In this context, the progress of AI applications is based on \begin{enumerate*}[label=(\arabic*)] \item using data generated or stored at the edges, \item working with large amounts of data from a wide range of sources, and \item protecting  data privacy in order to comply with the legal restrictions and to pay attention to end users' concerns.\end{enumerate*} Some use cases of AI with these dependencies are:




\begin{itemize}
    \item When data contains sensitive information, such as email accounts, personalised recommendations or health records, applications should employ privacy-preserving techniques to learn from a population of users whilst keeping the sensitive information on each user’s device \cite{Jalalirad2019ASA}.
   \item When information is located in data silos, for instance, healthcare industry is usually reluctant to disclose its records, keeping it as sequestered data \cite{bib:brisimi18}. Nevertheless, joint learning from data silos of different health institutions would allow to improve the robustness of the resulting models.
    \item Due to data privacy legislation, banks \cite{banks_app} and telecom \cite{8737464} companies cannot share individual records. However they would benefit from models that learn from several entities' data.
\end{itemize}

The standard machine learning paradigm does not match the previous dependencies, as it learns from a centralised data source. Likewise, distributed machine learning does not fit the preserving data privacy challenge, because data is shared among several computational elements. Moreover, distributed machine learning cannot cope with the  challenges associated to decentralised data processing, such as the ability to work with a great amount of clients with non homogeneous data distributions \cite{bib:konecny16federatedlearning}.

Federated learning (FL) is a nascent machine learning paradigm where many clients, in the sense of electronic devices or entire organisations, jointly train a model under the orchestration of a central server, while keeping the training data decentralised \cite{kairouz2019advances}. Roughly speaking, data is not shared with the central server, indeed it is kept in the devices where it is stored or generated. Accordingly, FL addresses the challenges of developing AI services on scattered data across a large amount of clients with non homogeneous data distributions.

Maintaining the data in its corresponding storage silos does not completely assure privacy preservation, since several adversarial attacks can still be damaging \cite{ji2019}. Data obfuscation, anonymisation techniques, such as blindly trusting artificial intelligence black box models (i.e. convolutional neural networks), or randomly sampling data from the clients' models have been proven to be inadequate to preserve privacy \citep{fredrikson2015model,chaudhuri2006random}. Moreover, the complete obfuscation of the data greatly reduces its value, thus a balance between privacy and utility is needed. Differential privacy (DP) is proposed as a data access technique which aims to maintain personal data privacy while maximising its utility \citep{TCS-042}.

The characteristics of FL and DP, and by extension their combination, make them candidates to address the challenges of distributed AI services that preserve data privacy. The research and progress of FL and DP need the support of software tools that ease the design of privacy-preserving AI services while not requiring development from scratch. Consequently, in recent years several software tools with FL and DP functionalities have been released with this aim.

We perform a comparative analysis of the FL and DP software tools released so far, and we conclude that their lack of a holistic view of FL and DP hinders the development of unified FL and DP AI services, as well as the furtherance of addressing the challenges of AI services at the edges that preserve data privacy. Therefore, we present \sherpa{the \texttt{Sherpa.ai} Federated Learning framework},\footnote{\url{https://developers.sherpa.ai/privacy-technology/}}\textsuperscript{,}\footnote{\url{https://github.com/sherpaai/Sherpa.ai-Federated-Learning-Framework}} an open-source \correcciones{unified} FL and DP framework for AI.

\sherpa{\texttt{Sherpa.ai} FL} aims to bridge the gap between the fundamental and applied research. Moreover, it will facilitate open research and development of new solutions built upon FL and DP for the challenges posed by AI at the edges and data privacy protection. A flexible approach to a wide range of problems is assured by its modular design that takes into account all the key elements and functionalities of FL and DP, which consist of:

\begin{enumerate}[noitemsep]
\item Data. Different data sets can be processed.
\item Learning model. Several core machine learning algorithms are incorporated.
\item Aggregation operator. Different operators for fusing the parameters of the clients' learning models are embodied.
\item Clients. It is where the learning models are run.
\item Federated server. The clients can be orchestrated by different communication strategies.
\item Communication among clients and server. Different solutions are encompassed to reduce the communication iterations, to protect the learning from adversarial attacks, and to obfuscate the parameters with DP techniques.
\item \correcciones{DP mechanisms. The fundamental DP mechanisms, such as the Laplace mechanism, as well as the composition of DP mechanisms are incorporated.}
\end{enumerate}

The progress of AI is not only supported by the release of software tools, but it needs fundamental guidelines defining how to put together the different software tools' attributes for reaching the intended learning goal while at same time matching the problem restrictions. Accordingly, since FL is a machine learning paradigm, we first study the principles of machine learning and how to make them fit the FL requirements. We see that most machine learning methods can be directly adapted to a FL setting, but some of them require  ad-hoc amendments. As a result of this study, we define the experimental workflow of FL in terms of methodological guidelines for preserving data privacy in the development of AI services at the edges. These methodological guidelines are grounded in the machine learning workflow, and they have guided the design and development of \sherpa{\texttt{Sherpa.ai} FL}, therefore they can be followed with \sherpa{\texttt{Sherpa.ai} FL}.

It is shown how to follow the mentioned methodological guidelines with \sherpa{\texttt{Sherpa.ai} FL} through two examples encompassing a classification and a regression use cases, namely:

\begin{enumerate}
    \item Classification. We use the EMNIST Digits dataset to describe how to conduct a classification task with \sherpa{\texttt{Sherpa.ai} FL}. We also compare the federated classification with its centralised counterpart. Both approaches achieve similar results.
    \item Regression. We describe how to perform a regression experiment using the California Housing dataset. We compare the FL experiment with its centralised version. In addition, it is shown how to assess and  limit the privacy loss using DP.
\end{enumerate}



The main contributions of this paper are:

\begin{enumerate}
    \item To analyse the most recently released FL and DP software tools, revealing the lack a unified view of FL and DP that hinders the possibility of addressing the challenges of AI at the edges with data privacy.
    \item To present \sherpa{\texttt{Sherpa.ai} FL},  an open-source unified FL and DP framework for AI.
    \item To study the adaptation of machine learning models to the principles of FL and, accordingly, to define the methodological guidelines which can be followed with \sherpa{\texttt{Sherpa.ai} FL} for developing AI services that preserve data privacy with FL and DP.
\end{enumerate}

The rest of the paper is organised as follows: the next section formally defines FL and DP as well as their key elements. Section \ref{sec:frameworks} analyses the main FL and DP frameworks' features. Section \ref{sec:software} introduces \sherpa{\texttt{Sherpa.ai} FL} including software architecture and functionalities. Section \ref{s_ml_fl} explains the adaptation of the machine learning paradigm to FL, taking into account the adaptation of core algorithms and the methodological guidelines of an experimental workflow. Section \ref{sec:illustrative-example} shows some illustrative examples consisting in a classification and regression problem. Finally, the concluding remarks and future work are reported in Section \ref{sec:conclusion}.


}

\vspace*{2.35em}

\section{Federated Learning and Differential Privacy}
\label{sec:federatedlearning}

\eugenio{The development of a framework for FL \correcciones{and DP} requires a thorough understanding of what FL is and what its key elements are. Accordingly, we formally define FL in Section \ref{definition}, and we detail each key element of a FL scheme in Section \ref{key-elements}.}
\correcciones{Similarly, DP is defined in Section \ref{sec:dp_definition}, and its key elements are described in Section \ref{sec:dp_key_elements}.
}

\subsection{The definition of Federated Learning}\label{definition}

\begin{figure*}[h!]
      \centering
      \includegraphics[width = \linewidth]{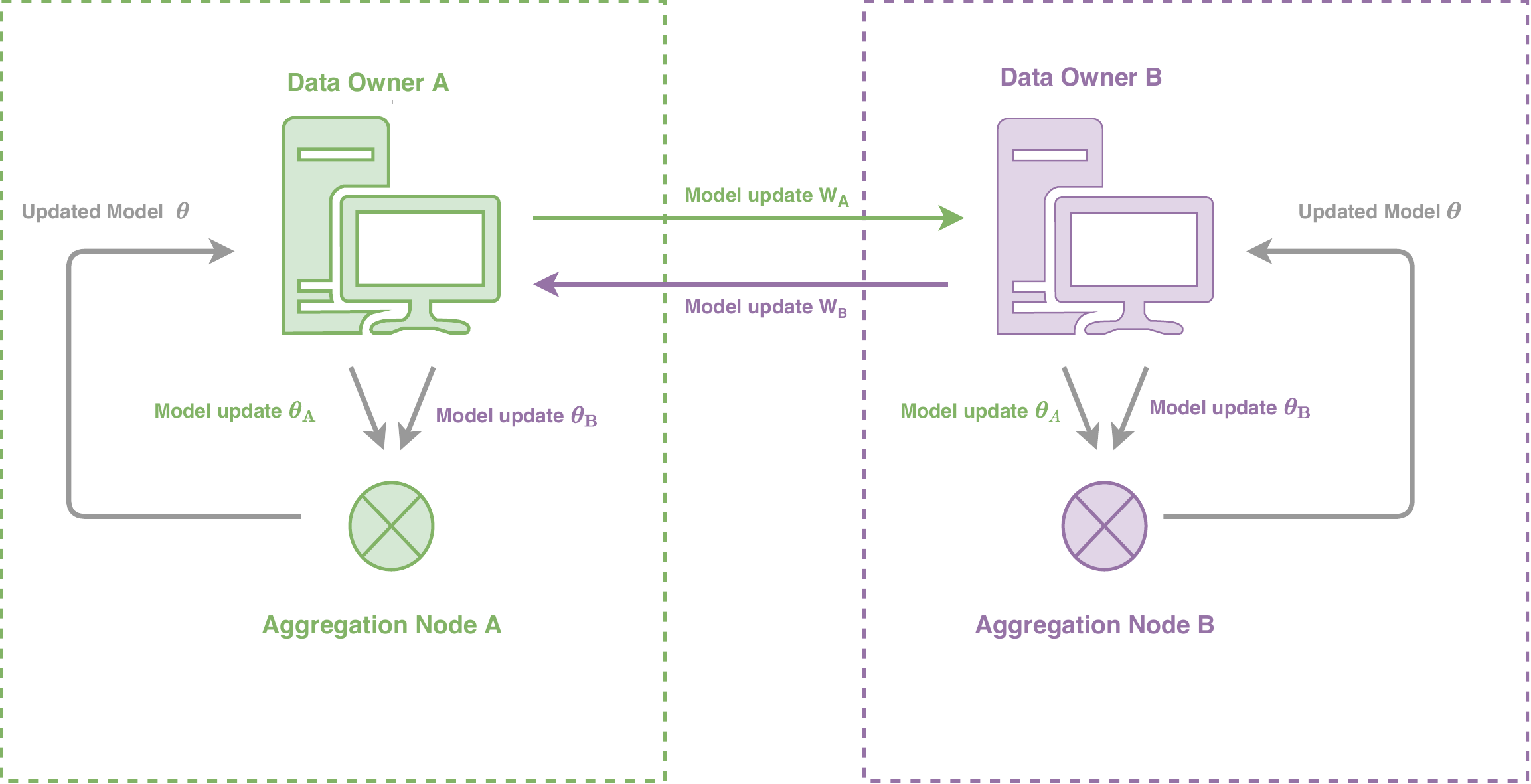}
      \caption{Representation of peer-to-peer FL architecture.}
      \label{peertopeer}
    \end{figure*}

\begin{figure*}[h!]
      \centering
      \includegraphics[width = 0.65\linewidth]{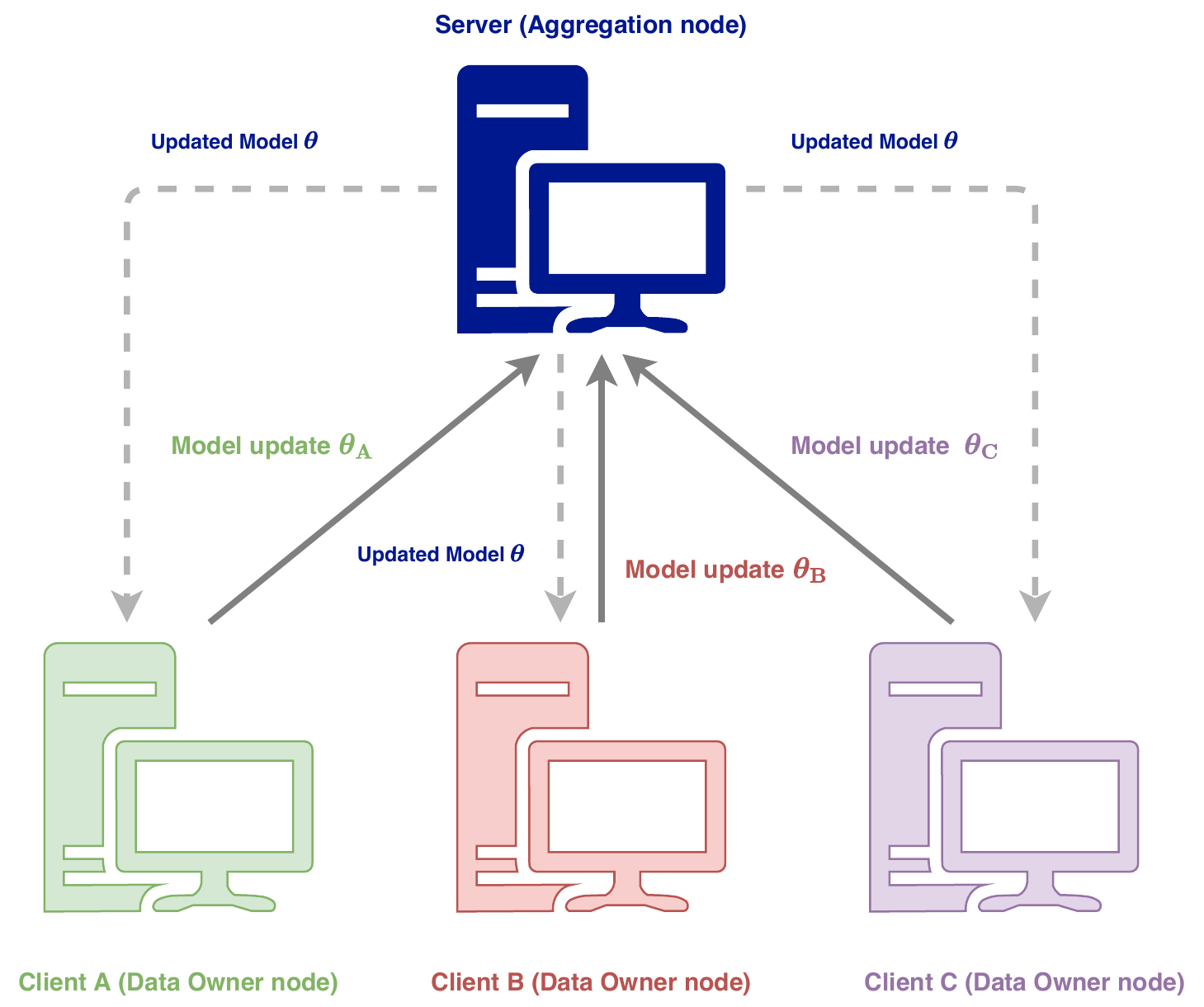}
      \caption{Representation of client-server FL architecture.}
      \label{client-server}
    \end{figure*}

FL is a distributed machine learning paradigm that consists of a network of nodes where we distinguish two types of nodes: \begin{enumerate*}[label={(\arabic*)}]\item \textit{Data owner} nodes, $\{C_1, \dots, C_n\}$, that possess a collection of data, $\{D_1, \dots, D_n\}$, and \item  \textit{Aggregation} nodes, $\{G_1, \dots, G_k\}$, aiming at learning a model from data owners.\end{enumerate*} The deployment of these two types of nodes defines, at least, two kind of federated architectures according to \citet{doi:10.2200/S00960ED2V01Y201910AIM043}, namely:

\nuria{\begin{enumerate}
    \item Peer-to-peer: It is the architecture in which all the nodes are both \textit{Data owner} and \textit{Aggregation} nodes. This scheme does not require a coordinator. The main advantages are the elevated security and data privacy while the main disadvantage is the computation cost. This FL architecture is illustrated in Figure \ref{peertopeer}.
    
    \item Client-server: It consists of a coordinator \textit{Aggregation} node named server and a set of \textit{Data owner} nodes named clients.\footnote{In the literature we find different ways to refer to the clients in a FL architecture, namely: nodes, agents or clients. In this paper, we rather prefer the term clients.} In this architecture, the client does not share its local data ensuring its privacy. We represent the client-server scheme in Figure \ref{client-server}.
\end{enumerate}}

Since the peer-to-peer model is a generalisation of the client-server model, we consider the latter for the formal definition of FL. In this architecture, each of the clients $C_i$ has a local learning model $LLM_i$ represented by the parameters $\theta_i$. FL aims at learning the global learning model $GLM$, represented by $\theta$, using the scattered data across clients through an iterative learning process known as \textit{round of learning}. For that purpose, in each round of learning $t$, each client trains the $LLM_i$ over its local training data $D^t_i$, updating their local parameters $\theta
^t_i$. Subsequently, the global parameters $\theta^t$ are computed aggregating the local parameters $\{\theta^t_1, \dots, \theta^t_n\}$ using a specific federated aggregation operator $\Delta$: 

\begin{equation}
    \theta^{t} = \Delta(\theta_1^{t},\theta_2^{t}, \dots, \theta_n^{t})
    \label{eq_fl_aggregation}
\end{equation}

After the aggregation of the parameters in the GLM, the LLMs are updated with the aggregated parameters:

\begin{equation}
    \theta_i^{t+1} \leftarrow \theta^t, \quad \forall i \in \{1, \dots, n\}
\end{equation}

The communication between server and clients can be synchronous or asynchronous. In the first option, the server awaits for the clients updates, aggregates all the local parameters and sends them to each client. Nevertheless, in the second option, the server merges the local parameters with the GLM as soon as it receives them, using a weighted scheme based on the age difference among the models.

We repeat this iterative process for as many rounds of learning as needed. Thus, the final value of $\theta$ will sum up the clients' underlying knowledge. 

In particular, the \correcciones{learning goal} is typically to minimise the following objective function:

\begin{equation}\label{eq_fl_objective_function}
    \min_\theta F(\theta) \textnormal{, \quad with} \quad F(\theta) := \sum_{i=1}^{n} w_i F_i(\theta)
\end{equation}

where $n$ is the number of clients, $F_i$ is the local objective function for the $i$-th client \correcciones{which is the common objective function of the problem fitted to each client's data}, $w_i \geq 0$ and $\sum_i w_i =1$.

\subsection{Key elements of Federated Learning}\label{key-elements}

The development of a FL environment requires the right combination of a set of necessary key elements. Since FL is a specific configuration of a machine learning environment, FL shares some key elements with it, namely: (1) data and (2) the learning model. However, the particularities of FL make necessary additional key elements, such as: (1) federated aggregation operators, (2) clients, (3) federated server and (4) communication among the federated server and the clients. The adaptation of the common key elements among FL and machine learning, and the FL specific ones are described as what follows.

\paragraph{\textbf{Data}}


Data plays a central role in FL as in machine learning. The distribution of data becomes crucial in FL since it is distributed among the different clients. \nuria{Regarding the splitting of the data among clients, there are two possibilities depending on the data distribution:


\begin{itemize}
    \item \textbf{IID (Independent and Identically Distributed) scenario:} when the data distribution in each client corresponds to the population data distribution. In other words, the data \correcciones{in each client} is independent and identically distributed, \correcciones{as well as} representative of the population data distribution.
    \item \textbf{Non-IID (non Independent and Identically Distributed) scenario:} when the data distribution in each client is not independent or identically distributed from the population data distribution.
\end{itemize}

In a real FL scenario, each client only stores the data generated on the client itself, ensuring the non-IID property of the global data. Hence, the non-IID scenario is the most likely one and it represents a real challenge for FL.}

\paragraph{\textbf{Learning model}}

The learning model is the shared structure between the server and the clients, where each client \correcciones{trains} a local model using its own data, while the global model on the server is never trained, but instead it is obtained aggregating the clients' model parameters. 
Thus multiple models are trained without explicit data sharing, a configuration that is essentially different from the classical (centralised) learning paradigm.

\paragraph{\textbf{Federated aggregation operators}}

The aggregation operator is in charge of  aggregating the parameters in the server. It has to: (1) assure a proper fusion of the local learning models in order to optimise the objective function in Equation \ref{eq_fl_objective_function}; (2) reduce the number of communication  rounds among the clients and  the federated  server and (3) be robust against clients with poor  data quality or malicious clients.

Some of the most commonly used federated aggregation operators in the literature are:

\begin{itemize}
    
    \item \textbf{Federated Averaging (FedAvg)} \cite{bib:mcmahan16communicationefficient}. It is based on keeping a shared global model that is periodically updated by averaging models that have been trained locally on clients. The training process is arranged by a central server which hosts the shared global model. However, the actual optimisation is done locally on clients. 
    
    
    \item \textbf{CO-OP} \cite{Wang2017COOPCM}. It proposes an asynchronous approach, which merges any received client model with the global model. Instead of directly averaging the models, the merging between a local model and the global model is carried out using a weighting scheme based on a measure of the difference in the age of the models. This is motivated by the fact that in an asynchronous framework, some clients will be trained on obsolete data while others will be trained on more up-to-date data.
\end{itemize}

\paragraph{\textbf{Clients}}

Each client of a federated scenario represents a node of the distributed scheme. Typical clients in FL could be smartphones, IoT devices or connected vehicles. Each client owns its specific training dataset and its local model. Their principal aim is to train local models on their own private data and share the trained model \ugr{parameters} with the federated server \correcciones{where the parameters fusion is performed}.

\paragraph{\textbf{Federated server}}

The federated server orchestrates the iterative learning of FL, which is composed of several rounds of learning. The server participates in: \begin{enumerate*}[label={(\arabic*)}] \item receiving the trained \ugr{parameters} of the local models, \item aggregating the trained \ugr{parameters} of each client model using federated aggregation operators and \item updating every learning model with the aggregated \ugr{parameters}. \end{enumerate*} The learning process involved in both training the local models and updating them, is known as a \correcciones{round of learning}. The global model, which is stored in the federated server, represents the final model after the learning process. Therefore it is used for predicting, testing or any posterior evaluation.

\paragraph{\textbf{Communication among the federated server and the clients}}
Communication between clients and server is the most tricky element of a FL scheme. On the one hand, an efficient communication is a crucial requirement due to the high communication times needed because of the network speed limitations and availability. For that reason, FL should minimise communications and maximise their efficiency by means of, for example, reducing the number of rounds of learning. On the other hand, the interchange of model \ugr{parameters} between the server and the clients constitutes a vulnerability to the federated server scheme, since \eugenio{the original data may be reconstructed from the model parameters \correcciones{through model-inversion adversarial attacks \cite{fredrikson2014}}, resulting in a great risk of private data leakage}. For this reason, DP techniques \cite{TCS-042} are commonly used in order to share model \ugr{parameters} \cite{mcmahan2018}.

\subsection{The definition of Differential Privacy}\label{sec:dp_definition}

DP is the property of an algorithm whose input is typically a database, and whose encoded response allows to obtain relatively accurate answers to potential queries \cite{TCS-042, Dwork2006}. 
The motivation for DP stems from the necessity of ensuring the privacy of individuals whose sensitive details are part of a database, while at the same time being able to gain accurate knowledge about the whole population when learning from the database.
DP does not imply a binary concept, \correcciones{\em{i.e.},} the guarantee or not of an individual's data privacy. 
Instead DP establishes a formal \textit{measure of privacy loss}, allowing for comparison between different approaches. \correcciones{Thus, DP will rigorously bound the possible harm to an individual whose sensitive information belongs to the database by fixing a budget for privacy loss.}

The formal definition of DP requires a few preliminary notions. Namely, we define the \textit{probability simplex} over a discrete set $B$, denoted $\Delta (B) $, as the set of real valued vectors whose $|B|$ components sum up to one and are non-negative: 
\begin{equation}
	\Delta (B) := \left\{ 
		x \in \mathbb{R}^{|B|} \,:\, \sum_{i=1}^{|B|} x_i = 1,\, 
		x_i \geq 0,\, i=0,\dots,|B|
		\right\}
	\label{eq_dp_probability_simplex}
\end{equation}

A \textit{randomised algorithm} $\mathcal{M}: A \rightarrow B$, with $B$ a discrete set, is defined as a mechanism which is associated with a mapping $M: A \rightarrow \Delta( B )$ such that, with input $a \in A$, the mechanism produces $\mathcal{M}(a)= b$ with probability $(M(a))_b = P(b|a)$, for each $b \in B$. 
The probability is taken over the randomness employed by the mechanism $\mathcal{M}$.   

In general, databases are collections of records from a universe $\mathcal{X}$. It is convenient to express databases $x$ by their histogram $x \in \mathbb{N}^{|\mathcal{X}|}$, where each component $x_i$ stands for the number of elements in the database of \textit{type i} in $\mathcal{X}$. 

This interpretation naturally leads to define the \textit{distance between databases}: two databases $x,\,y$  are said to be $n$-neighbouring if they differ by $n$ entries as $||x-y||_1 = n$, where $||\cdot||_1$ denotes the $\ell_1$ norm. In particular, if the databases only differ in a single data element ($n=1$), the databases are simply addressed as \textit{neighbouring}. 

At this stage, \textit{DP} can be formally introduced. 
A randomised algorithm (mechanism) $\mathcal{M}$ with domain $\mathbb{N}^{|\mathcal{X}|}$ preserves $\epsilon$-DP for $\epsilon > 0$ if for all neighbouring databases $x,\,y \in \mathbb{N}^{|\mathcal{X}|}$ and all $ \mathcal{S} \subseteq \textnormal{Range}(\mathcal{M})$ it holds that:
\begin{equation}\label{eq_dp_definition}
	P[\mathcal{M}(x) \in \mathcal{S}] \leq \exp (\epsilon)  P [\mathcal{M}(y) \in \mathcal{S}] 
\end{equation}
If, on the other hand, for $0<\delta< 1$ it holds that:
\begin{equation}\label{eq_dp_weaker_definition}
P[\mathcal{M}(x) \in \mathcal{S}] \leq \exp (\epsilon)  P [\mathcal{M}(y) \in \mathcal{S}] + \delta
\end{equation}
then the mechanism possesses the \textit{weaker} property of $(\epsilon, \delta)$-DP. 
The probability is taken over the randomness employed by the mechanism $\mathcal{M}$.


In essence, Equation \ref{eq_dp_definition} tells us that for every run of the randomisation mechanism $\mathcal{M}(x)$, it is almost equally likely to observe the same output for \textit{every} neighbouring database $y$, such probability is governed by $\epsilon$. Equation \ref{eq_dp_weaker_definition} is weaker since it allows us to exceed $\epsilon$ with probability $\delta$. 

In other words, DP specifies a ``privacy budget'' given by $\epsilon$ and $\delta$. The way in which it is spent is given by the concept of privacy loss. We define the privacy loss incurred in observing the output $\mu$ employing the randomised algorithm $\mathcal{M}$ in two neighbouring databases $x, y$:
\begin{equation} \label{eq_priv_loss}
    \mathcal{L}^{\mu}_{\mathcal{M}(x)|| \mathcal{M}(y)} := \ln \bigg( \frac{P[\mathcal{M}(x)=\mu]}{P[\mathcal{M}(y)=\mu]} \bigg)
\end{equation}
%

Since the privacy loss can be both positive and negative, we consider the absolute value of it in the following interpretation. The privacy loss allows us to reinterpret both $\epsilon$ and $\delta$ in a more intuitive way:

\begin{itemize}
    \item $\epsilon$ limits the quantity of privacy loss permitted, that is, our privacy budget. 
    \item $\delta$ is the probability of exceeding the privacy budget given by $\epsilon$, so that we can ensure that with probability $1-\delta$, the privacy loss will not be greater than $\epsilon$.
\end{itemize}


DP is immune to post-processing, that is, if and algorithm protects an individual's privacy, then there is not any way in which privacy loss can be increased, stated in a more formal way: let $\mathcal{M}: \mathbb{N}^{|\mathcal{X}|} \rightarrow \mathcal{R}$ be a ($\epsilon, \delta$)-differentially private mechanism and let $f: \mathcal{R} \rightarrow \mathcal{R}'$, then $f \circ \mathcal{M}: \mathbb{N}^{|\mathcal{X}|} \rightarrow \mathcal{R}'$ is ($\epsilon, \delta$)-differentially private.

\subsection{Key elements of Differential Privacy} \label{sec:dp_key_elements}


DP arose as the principal setting for privacy-preserving sensitive data when delivering trained models to \correcciones{untrusted} parties. \correcciones{The possibilities of DP are built upon the modular structure of its elements, which allows to construct more sophisticated DP mechanisms, and to design, analyse and post-process DP mechanisms for a specific privacy-preserving learner \cite{TCS-042, Rubinstein2017}. These necessary or key elements of DP are the DP mechanisms, the composition DP mechanisms, and the subsampling techniques to increase the privacy. We subsequently detail them.}


\paragraph{\textbf{DP mechanisms}} 

\correcciones{We describe the main privacy-preserving mechanisms as what follows:}

\begin{itemize}
\item \textbf{Randomised response mechanism}.
\correcciones{It} is aimed at evaluating the frequency of an embarrassing or illegal practice. When answering whether it engaged in the aforementioned activity in the past period of time,  the following procedure is proposed: 
\begin{enumerate}
	\item Flip a coin;
	\item If tails,  respond truthfully;
	\item If heads,  flip a second coin and if heads, respond ``Yes'',  and respond ``No'' if tails.
\end{enumerate}
This approach provides privacy due to ``plausible deniability'' since the response ``Yes'' may have been submitted when both coins flips turned out heads. 
By direct computation it can be shown that this is an $\epsilon$-differentially private mechanism with $\epsilon = \log(3)$  \cite[Section 3.2]{TCS-042}.    
\item \textbf{Laplace mechanism} \citep{Dwork2006}.
\correcciones{It} is usually employed for preserving privacy in numeric queries $f: \mathbb{N}^{|\mathcal{X}|} \rightarrow \mathbb{R}^k$, which map databases $x\in \mathbb{N}^{|\mathcal{X}|}$ to $k$ real numbers.
At this point, it is important to introduce a key parameter associated to the \textit{accuracy} of such queries, namely the $\ell_1$ \textit{sensitivity}:
\begin{equation}\label{eq_dp_l1_sensitivity}
	\Delta f := \underset{||x-y||_1 = 1 }
	\max \left\Vert f(x) - f(y)\right\Vert_1
\end{equation}
Since the above definition must hold for every neighbouring $x,y \in \mathbb{N}^{|\mathcal{X}|}$, it is also denoted as \textit{global sensitivity} \cite{Rubinstein2017}.
This parameter measures the maximum  magnitude of change in the output of $f$ associated to a single data element, thus, intuitively, it establishes the amount of uncertainty (\correcciones{\em{i.e.},} noise) to be introduced in the output to preserve the privacy of a single individual. 

Moreover, we denote as Lap$(b)$ the Laplace distribution with  probability density function  with scale $b$ and centred at $0$.  Given any function $f: \mathbb{N}^{|\mathcal{X}|} \rightarrow \mathbb{R}^k$, the Laplace mechanism can be  defined as 
\begin{equation}\label{eq_dp_laplace_mechanism}
	\mathcal{M}_L(x, f(\cdot), \epsilon) := f(x) + \left(Y_1,\dots, Y_k \right)
\end{equation} 
where the components $Y_i$ are IID drawn from the distribution Lap$(\Delta f/\epsilon)$.
In other words, each component of the output of $f$ is perturbed by Laplace noise according to the \correcciones{sensitivity of the function} $\Delta f$.
It can be shown that this is an $\epsilon$-differentially private mechanism with $\epsilon = \Delta f /b$  \cite[Section 3.3]{TCS-042}.

\item \textbf{Exponential mechanism} \citep{McSherry2007}.
\correcciones{It} is a \textit{general} DP mechanism that has been proposed for situations in which adding noise directly to the output function (as for Laplace mechanism) would completely ruin the result. 
Thus the exponential mechanism constitutes the building component for queries with arbitrary utility, where the goal is to \textit{maximise} the utility while preserving privacy. 
For a given arbitrary range $\mathcal{R}$, the utility function $u:\mathbb{N}^{|\mathcal{X}|} \times \mathcal{R} \rightarrow \mathbb{R}$ maps database/output pairs to utility values. 
We introduce the \textit{sensitivity of the utility function} as 
\begin{equation}\label{eq_dp_utility_sensitivity}
	\Delta u := \underset{r\in\mathcal{R}}\max \; \underset{||x-y||_1 \le 1}\max |u(x,r) - u(y,r)|
\end{equation}
where the sensitivity of $u$ with respect to the database is of importance, while it can be arbitrarily sensitive with respect to the range $r \in \mathcal{R}$.   
The exponential mechanism $\mathcal{M}_E(x,u,\mathcal{R})$ is defined as a randomised algorithm which picks as output an element of the range $r \in \mathcal{R}$ with probability proportional to $\exp\left( \epsilon u(x,r) / (2\Delta u) \right)$.

When normalised, the mechanism details a probability density function over the possible responses $r \in \mathcal{R}$.
Nevertheless, the resulting distribution can be rather complex and over an arbitrarily large domain, thus the implementation of such mechanism might not always be efficient \cite{TCS-042}.
It can be shown that this is a $(2\epsilon\Delta u)$-differentially private mechanism \cite{McSherry2007}.


\item \textbf{Gaussian mechanism} \citep{TCS-042}.
\correcciones{It} is a DP mechanism that adds Gaussian noise to the output of a numeric query. It has two great advantages over the differentially private mechanisms stated previously:
\begin{itemize}
    \item \textbf{Common source noise}: the added Gaussian noise is the same as the one which naturally appears when dealing with a database.
    \item \textbf{Additive noise}: the sum of two Gaussian distributions is a new Gaussian distribution, therefore it is easier to statistically analyse this DP mechanism.
\end{itemize}
Instead of scaling the noise to the $\ell_1$ \textit{sensitivity}, as we previously did with the Laplacian mechanism, it is scaled to the $\ell_2$ \textit{sensitivity}:
\begin{equation}\label{eq_dp_l2_sensitivity}
	\Delta_2 (f) := \underset{||x-y||_1 = 1 }
	\max \left\Vert f(x) - f(y)\right\Vert_2
\end{equation}

Moreover, we denote as N$(0,\sigma^2)$ the Gaussian distribution with probability density function  with mean $0$ and variance $\sigma^2$.  Given any function $f: \mathbb{N}^{|\mathcal{X}|} \rightarrow \mathbb{R}^k$, the Gaussian mechanism can be defined as:
\begin{equation}\label{eq_dp_gaussian_mechanism}
	\mathcal{M}_G(x, f(\cdot), \epsilon) := f(x) + \left(Y_1,\dots, Y_k \right)
\end{equation} 
where the components $Y_i$ are IID drawn from the distribution N$(0,\sigma)$.

However, it needs to satisfy the following restrictions to ensure it is a ($\epsilon, \delta$)-differentially private mechanism: for $\epsilon \in (0,1)$ and variance $\sigma^2 > 2 \cdot \ln(1.25/\delta) \cdot (\Delta_2(f) / \epsilon)^2$, the Gaussian mechanism is ($\epsilon, \delta$)-differentially private.

\end{itemize}

To sum up, the main idea behind DP mechanisms is adding a certain amount of noise to the query output, while preserving the utility of the original data. Such noise is calibrated to the privacy parameters $(\epsilon, \delta)$ and the sensitivity of the query function. 

\paragraph{\textbf{Composition of DP mechanisms}}

An appealing property of DP is that more advanced private mechanisms can be devised by combining DP mechanisms, such as the general building components described in Section \ref{sec:dp_key_elements}. The resulting  mechanism then still preserve DP, and the new values of $\epsilon$ and $\delta$ can be computed according to the composition theorems. Before the composition theorems are provided, we state an experiment with an adversarial which proposes a composition scenario for DP \citep{TCS-042}.

\paragraph{Composition experiment $b \in \{ 0,1 \}$ for adversary $A$ with a given set, $M$, of DP mechanisms}

For $i=1,\dots,k$:

\begin{enumerate}
    \item $A$ generates two neighbouring databases $x_i^0$ and $x_i^1$ and selects a mechanism $\mathcal{M}_i$ from $M$.
    \item $A$ receives the output $y_i \in \mathcal{M}_i(x_i^b)$
\end{enumerate}

In the experiments the adversary preserves its state between iterations, and we define $A$'s view of the experiment $b$ as $V^b=\{y_1, \dots, y_k\}$. In order to ensure DP in these Composition experiments we need to introduce a statistical distance which resembles the privacy loss (Equation \ref{eq_priv_loss}). 

The $\delta$-Approximate Max Divergence between random variables Y and Z is defined as:
\begin{equation}
    D_{\infty}^{\delta}(Y||Z) = \max_{P[Y\in S] > \delta} \ln \frac{P[Y\in S] - \delta}{P[Z \in S]}
\end{equation}

We say that the composition of a sequence of DP mechanisms under the Composition experiment is ($\epsilon, \delta$)-differentially private if $D_{\infty}^{\delta} ( V^0 || V^1 ) \leq \epsilon$. Now, we are ready to introduce the composition theorems:

\begin{itemize}

\item \textbf{Basic composition theorem}. \label{basic_comp} The composition of a sequence $\{\mathcal{M}_k\}$ of ($\epsilon_i, \delta_i$)-differentially private mechanisms under the Composition experiment with $M=\{\mathcal{M}_k\}$, is ($\sum_{i=1}^{k} \epsilon_i, \sum_{i=1}^{k} \delta_i$)-differentially private.

\item \textbf{Advanced composition theorem}. \label{advanced_comp} For all $\epsilon, \delta, \delta' \geq 0$ the composition of a sequence $\{\mathcal{M}_k\}$ of ($\epsilon, \delta$)-differentially private mechanisms under the Composition experiment with $M=\{\mathcal{M}_k\}$, satisfies ($\epsilon', \delta''$)-DP with:
\begin{equation} \label{eq_advanced_comp}
\epsilon' = \epsilon \sqrt{2k\ln(1/\delta')} + k \epsilon(e^{\epsilon}-1) \quad \textnormal{and} \quad \delta'' = k\delta + \delta'
\end{equation}

More advanced versions of Equation \ref{eq_advanced_comp} that allow the composition of private mechanisms with diverse $\epsilon$ and $\delta$ values and provide tighter bounds can be found in \citep{bib:kairouz17}.

\end{itemize}

\paragraph{\textbf{Privacy filters} \citep{privacyfilters}} 
While composition theorems are quite useful, they require some parameters to be defined upfront, such as the number of mechanisms to be composed. Therefore, no intermediate result can be observed and the privacy budget can be wasted. In such situations it is required a more fine grained composition techniques which allows to observe the result of each mechanism without compromising the privacy budget spent.

In order to remove some of the stated constraints, a more flexible experiment of composition is introduced \citep{privacyfilters}:

\paragraph{Adaptive composition experiment $b \in \{ 0,1 \}$ for adversary $A$}

For $i=1,\dots,k$:

\begin{enumerate}
    \item $A$ generates two neighbouring databases $x_i^0$ and $x_i^1$ and selects a mechanism $\mathcal{M}_i$ that is ($\epsilon_i, \delta_i$)-differentially private.
    \item $A$ receives the output $y_i \in \mathcal{M}_i(x_i^b)$
\end{enumerate}

In these situations, the $\epsilon_i$ and $\delta_i$ of each mechanism is adaptively selected based on the outputs of previous iterations. For the adaptive composition experiment, the privacy loss of the adversary's view $V=\{y_1,\dots, y_k\}$ for each pair of neighbouring databases $x,y$ is defined as follows:
\begin{equation} \label{eq_priv_filter_loss}
    \mathcal{L}^{V} = \ln \bigg( 
        \frac{ \prod_{i=1}^k P[\mathcal{M}_i(x)=y_i|V_{i}] }
        { \prod_{i=1}^k P[\mathcal{M}_i(y)=y_i|V_{i}]} 
    \bigg)
\end{equation}
where we write $V_i=\{ y_1, \dots, y_i\}$, that is, the adversary's view at the beginning of the $i^{th}$-iteration of the adaptive composition experiment. In particular, if the adaptive composition experiment has only one iteration ($k=1$), the Equation \ref{eq_priv_filter_loss} is the same as the definition of privacy loss (see Equation \ref{eq_priv_loss}).

The function $ COMP_{\epsilon_g, \delta_g}: \mathds{R}_{\geq 0}^{2k} \rightarrow \{HALT, CONT\} $ is a valid privacy filter for $\epsilon, \delta \geq 0$ if for all adversaries in the adaptive composition experiment, the following "bad event" occurs with probability at most $\delta_g$ when the adversary's view V:

\begin{equation}
    |\mathcal{L}^{V}| > \epsilon_g \quad \textnormal{and} \quad COMP_{\epsilon_g, \delta_g}(\epsilon_1, \delta_1, \dots, \epsilon_k, \delta_k) = CONT
\end{equation}

A privacy filter can be used to guarantee that with probability $1-\delta_g$, the stated privacy budget $\epsilon_g$ is never exceeded. That is, fixed a privacy budget ($\epsilon_g, \delta_g$), the function $ COMP_{\epsilon_g, \delta_g}: \mathds{R}_{\geq 0}^{2k} \rightarrow \{HALT, CONT\} $ controls the composition. It returns HALT if the composition of $k$ given DP mechanisms surpasses the privacy budget, otherwise it returns CONT. Privacy filters have similar composition theorems to the ones given \correcciones{above}: 

\begin{itemize}
    \item \textbf{Basic composition for privacy filters}.
For any $\epsilon_g, \delta_g \geq 0,\ $ $COMP_{\epsilon_g, \delta_g}$ is valid Privacy Filter, where: 
\begin{equation*}
 COMP_{\epsilon_g,\delta_g}(\epsilon_1,\delta_1,...,\epsilon_{k},\delta_{k})= \begin{cases} 
      HALT & \textnormal{if}\  \sum_{i=1}^{k} \delta_i > \delta_g \ \ \ \textnormal{or} \ \ \ \sum_{i=1}^{k} \epsilon_i > \epsilon_g, \\
      CONT & \textnormal{otherwise}
\end{cases}
\end{equation*}

\item \textbf{Advanced composition for privacy filters}.
We define $\mathcal{K}$ as follows:
\begin{equation*}
    \mathcal{K} :=  
    \sqrt{\bigg( \sum_{i=1}^{k} \epsilon_i^2 + H \bigg) \bigg( 2 + \ln{\big( \frac{1}{H} \sum_{i=1}^{k} \epsilon_i^2 +1 \big)} \bigg) \ln{(2/\delta_g)}}
                    +
    \sum_{j=1}^{k} \epsilon_j \bigg( \frac{\exp{(\epsilon_j)}-1}{2} \bigg)
\end{equation*}

with $ H = \frac{\epsilon_g^2}{28.04 \ln(1/\delta_g)}$.\\
 
Then $COMP_{\epsilon_g, \delta_g}$ is a valid Privacy Filter for $\delta_g \in (0, 1/e)$ and $\epsilon_g > 0$, where:
\begin{equation*}
 COMP_{\epsilon_g,\delta_g}(\epsilon_1,\delta_1,...,\epsilon_{k},\delta_{k})= \begin{cases} 
      HALT & \textnormal{if}\  \sum_{i=1}^{k} \delta_i > \delta_g/2 \ \ \ \textnormal{or} \ \ \ \mathcal{K} > \epsilon_g, \\
      CONT & \textnormal{otherwise}
\end{cases}
\end{equation*}

\end{itemize}
The value of $\mathcal{K}$ might be strange at first sight, however if we assume $\epsilon_j=\epsilon$ for all $j$, it remains:
\begin{equation*}
    \mathcal{K} = \sqrt{ \bigg(k\epsilon^2 + H\bigg)\bigg(2+\ln{\left(\frac{k\epsilon^2}{H} + 1\right)}\bigg) \ln{(2/\delta)}} + k\epsilon^2 \left(\frac{\exp{(\epsilon)}-1}{2}\right)
\end{equation*}
which is quite similar to Equation \ref{eq_advanced_comp}.


\paragraph{\textbf{Increase privacy by subsampling}} \correcciones{The privacy of a DP mechanism can be further improved whether} instead of querying all the stored data, a random subsample is queried. That is, if an ($\epsilon, \delta$)-differentially private mechanism is used to query random subsample from a database with $n$ records, then an improved ($\epsilon', \delta'$) parameters can be provided according to the type of random subsample \citep{balle2018privacy}. If the random subsample of size $m<n$ is performed without replacement then:

\begin{equation} \label{ep_with_replacement}
    \epsilon' = \ln \left(1 + \frac{m}{n}\left(e^{\epsilon}-1\right)\right) \quad \text{and} \quad \delta' = \frac{m}{n} \delta
\end{equation}

This expression for ($\epsilon', \delta'$) is better than the original ($\epsilon, \delta$) in the sense that it is smaller and so is the privacy budget spent. The noise considered in such situation comes from a different source than the noise added by the DP mechanism itself. That is, the DP mechanism is adding a certain quantity of noise specified by the ($\epsilon, \delta$) parameters to a random subsample of the database, therefore the information extracted is influenced by the individuals contained in it. This random subsample is sampled each time the DP mechanism is used, which may result in slightly different results for the same query applied multiple times, that is, a new source of noise is added to the query. 

Particularly, the improvement is greatly noticeable when $\epsilon < 1$, which makes the Gaussian Mechanism ideal, since to achieve ($\epsilon, \delta$)-DP $\epsilon$ must be smaller than 1. That is, the Gaussian Mechanism and the subsampling methods, when applied together, can ensure a minor quantity of noise and a tinier privacy budget expenditure at the cost of accessing a small random subsampling of the data.

This technique is particularly suited for FL, where the data does not come from all the clients in each iteration, but it does from a random sample of them. Moreover, it is well suited for programs in which the privacy parameters are hardcoded, so the privacy budget must be carefully spent.

\section{ \correcciones{Software tools: FL and DP frameworks analysis}}
\label{sec:frameworks}

\correcciones{The high demand of AI services at the edges which must preserve data privacy has pushed the release of several software tools or frameworks of FL and DP. In this Section, we discuss the strengths and weaknesses of these software frameworks, we compare them and stress out their main shortcomings.\footnote{\correcciones{The discussion covers the state of the development of the software tools until the end of May 2020.}}
}



\subsection{PySyft}

PySyft\footnote{\url{https://github.com/OpenMined/PySyft}} is a Python library for secure and private deep learning. PySyft decouples private data from model training, using FL, DP, and Encrypted Computation (like Multi-Party Computation (MPC) and Homomorphic Encryption (HE)) within the main deep learning frameworks like PyTorch and TensorFlow.


\paragraph{\textbf{Features}}
It is compatible with existing deep learning frameworks such as TensorFlow and PyTorch. Their low level FL implementation allows developing and debugging projects with complex communication networks in a local environment with almost no overhead. It is mainly focused on providing Secure MPC through HE, it thus allows to apply computations on ciphertext which is ideal for developing FL models while preserving privately the results of the computations to the participants. Last, they offer many Python notebooks, which greatly softens the learning curve of this framework.

\paragraph{\textbf{Shortcomings}}
Its low level of FL support is missing some key features: neither it includes any dataset by default nor it implements any model aggregation operators. Its low level implementation and the two drawbacks stated before make this framework quite complex to use, requiring considerable knowledge in this field to correctly assemble a FL model.

While its webpage\footnote{\url{https://www.openmined.org}} advertises many DP mechanisms, they are nowhere to be found. As a matter of fact, in their github documentation they state the following: ``Do NOT use this code to protect data (private or otherwise) - at present it is very insecure. Come back in a couple of months''.

\paragraph{\textbf{Overview}} 
We conclude that PySyft is a low level FL framework for advanced users which is compatible with many well-known deep learning frameworks and it does provide neither any DP mechanism nor any DP algorithm.

\subsection{TensorFlow}

TensorFlow implements DP and FL through its libraries TensorFlow Privacy and TensorFlow Federated, respectively.

\paragraph{\textbf{Features}}

TensorFlow Privacy\footnote{\url{https://github.com/tensorflow/privacy}} is a Python library for training machine learning models with privacy for training data. It integrates seamlessly with existing TensorFlow models and allows the developer to train its models with DP techniques. In addition they have many tutorials to quickly learn how to use it. 

TensorFlow Federated\footnote{\url{https://www.tensorflow.org/federated}} is an open-source framework for machine learning and other computations on decentralised data. As TensorFlow Privacy, it integrates easily with existing TensorFlow Models. In addition, it has built-in many known training datasets. 

\paragraph{\textbf{Shortcomings}}

TensorFlow Privacy only focuses on differentially private optimisers and it does not provide any DP mechanisms to implement your own differentially private optimisers. It does not officially support any other deep learning library and it is still not compatible with the latest TensorFlow 2.x. In addition, it is a ''library under continual development'' according to its Github documentation\footnote{\url{https://github.com/tensorflow/privacy}}, it is not thus mature enough for production usage.

While TensorFlow Federated provides both low level and high level interfaces for FL settings and it has some high level interfaces to create aggregation operators, it does not provide any built-in aggregation operators. Last, it is not yet compatible with the latest TensorFlow 2.x.

\paragraph{\textbf{Overview}} 
These TensorFlow frameworks in conjunction allow us to develop FL models, but they are tied to the TensorFlow framework, which greatly denies any portability of the generated model. They are neither compatible with the latest version of TensorFlow nor they are ready for final products. In addition, they lack  DP mechanisms to implement new privacy-preserving algorithms.

\subsection{FATE}

FATE\footnote{\url{https://fate.fedai.org/overview/}} is an open-source project initiated by Webank's AI Department to provide a secure computing framework to support the federated AI ecosystem. 

\paragraph{\textbf{Features}}

It provides many interesting FL algorithms and it exposes\footnote{\url{ https://fate.readthedocs.io/en/latest/examples/federatedml-1.x-examples/README.html}} a high level interface driven by custom scripts.

\paragraph{\textbf{Shortcomings}}
Its high level interface made of scripts relies too much on command line parameters and on a poorly documented domain specific language. It is unclear how to implement a low level FL model, which makes us think this framework is designed as a black box model. Their modular architecture seems quite complex. Also, it does not feature any DP algorithm, and there are no signs of future plans for implementing them.

\paragraph{\textbf{Overview}} 
This framework is mainly focused on FL, making one of its biggest weaknesses that it does not implement any DP algorithm, in order to improve its data protection regulation compliance. Secure computation protocols ensure that data is not eavesdropped by an adversary, but it does not ensure that individuals' privacy, roughly speaking, is preserved. In addition, it is expected to be used as a high level interface which relies on a barely documented custom language.

\subsection{LEAF}

LEAF\footnote{\url{https://leaf.cmu.edu/}} is a benchmarking framework for learning in federated settings, with applications including FL, multi-task learning, meta-learning, and on-device learning.

\paragraph{\textbf{Features}}
This framework mainly focuses on benchmarking FL settings. It provides some basic FL mechanisms such as the Federated Averaging Aggregator and given its modular design it can be adapted to work on any existing framework. Last, it has some known built-in datasets such as FEMNIST, Shakespeare and Celeba.

\paragraph{\textbf{Shortcomings}}
It does not provide any benchmark for preserving privacy in a FL setting, even though privacy must be taken into consideration as it is a desired property of many FL settings. Moreover, it does not offer as many official documentation or tutorials as the other frameworks discussed in this section.

\paragraph{\textbf{Overview}}
LEAF offers a baseline implementation for some basic FL methods but its main purpose is benchmarking FL settings. However, DP benchmarks are not provided, even though nowadays privacy is a concern in most FL settings.

\subsection{PaddleFL} 

PaddleFL\footnote{\url{https://paddlefl.readthedocs.io/en/latest}} is an open source FL framework based on PaddlePaddle\footnote{\url{https://github.com/paddlepaddle/paddle}}. PaddlePaddle is an industrial platform with advanced technologies and rich features that cover core deep learning frameworks, basic model libraries, end-to-end development kits, tool and component as well as service platforms.

\paragraph{\textbf{Features}}

PaddleFL provides a high level interface to develop FL models with DP. In the FL field it implements the Federated Averaging Aggregator and its secure multi-party computation equivalent. When it comes to DP, it provides an implementation of the differentially private stochastic gradient descent.

\paragraph{\textbf{Shortcomings}}

This framework has little documentation. It lacks any other DP algorithm so there is great difficulty in developing alternative privacy-preserving techniques. Last, since it is based on PaddlePaddle it is not compatible with other frameworks, and there is  little documentation which  makes it really hard to use and understand.

\paragraph{\textbf{Overview}} 

PaddleFL provides a high level interface for some basic and well-known FL aggregators and implements a differentially private algorithm, being one of its main drawbacks that it is little documented and it does not implement any tool to easily extend its capabilities.

\subsection{\correcciones{Frameworks analysis}}

\correcciones{The discussed software tools share some shortcomings for developing distributed AI services that preserves data privacy. Among them, we stress out the following:}

\begin{enumerate}
    \item \correcciones{They focus on FL or DP, but they do not provide a unified approach for both of them.}
    \item \correcciones{They lack  DP mechanisms and related methods from the DP area. Likewise, they do not allow to develop and integrate new DP mechanisms in the frameworks.}
    \item Only the most basic federated \correcciones{aggregation} operators are implemented.
    \correcciones{They are mainly focused on deep learning models, and they do not provide support for other machine learning algorithms that may be also used in the FL setting.}
\end{enumerate}

\correcciones{We summarise and compare the characteristics of the frameworks reviewed in Table \ref{tab:comparison_without_sherpa}. We conclude that a unified FL and DP framework is required, and this is the ambitious aim of \sherpa{\texttt{Sherpa.ai} FL}, which we present in the following section.}


\begin{table}[!h]
\centering
\captionsetup{justification=centering,margin=0.5cm}
\resizebox{\textwidth}{!}{%
\begin{tabular}{|l|c|c|c|c|c|}
\rowcolor[HTML]{3b8faa} 
{\color[HTML]{FFFFFF} \textbf{\begin{tabular}[c]{@{}l@{}}FL \& DP features\end{tabular}}} &
  \makecell{\color[HTML]{FFFFFF}  \textbf{PySyft} \\ \includegraphics[width =  0.02\textwidth]{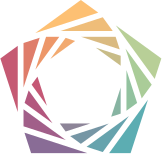}} &
  \makecell{\color[HTML]{FFFFFF}  \textbf{TensorFlow} \\ \includegraphics[width =  0.03\textwidth]{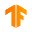}} &
  \makecell{\color[HTML]{FFFFFF} \includegraphics[width = 0.05\textwidth]{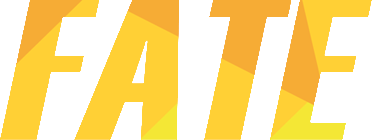}}  &
  \makecell{\color[HTML]{FFFFFF}  \textbf{LEAF} \\ \includegraphics[width = 0.02\textwidth]{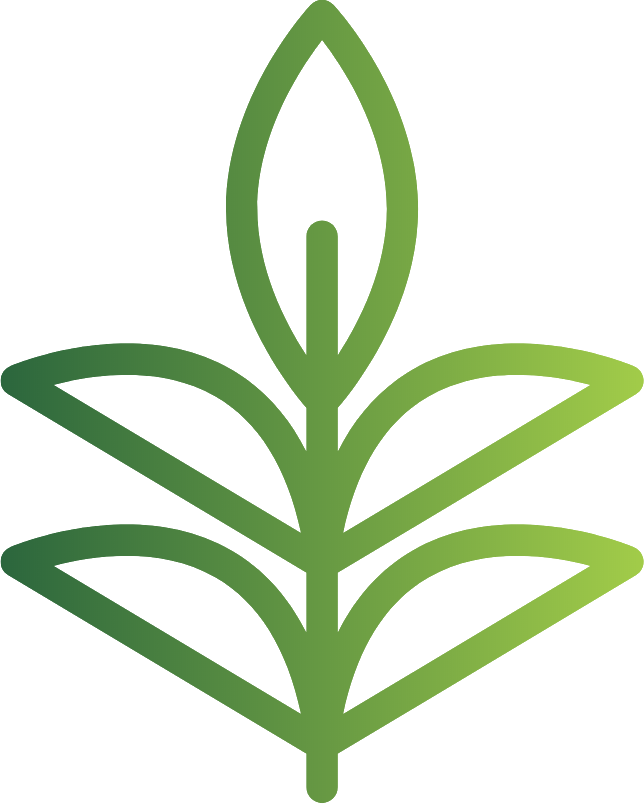}}  &
  \makecell{\color[HTML]{FFFFFF}  \textbf{PaddleFL} \\ \includegraphics[width = 0.02\textwidth]{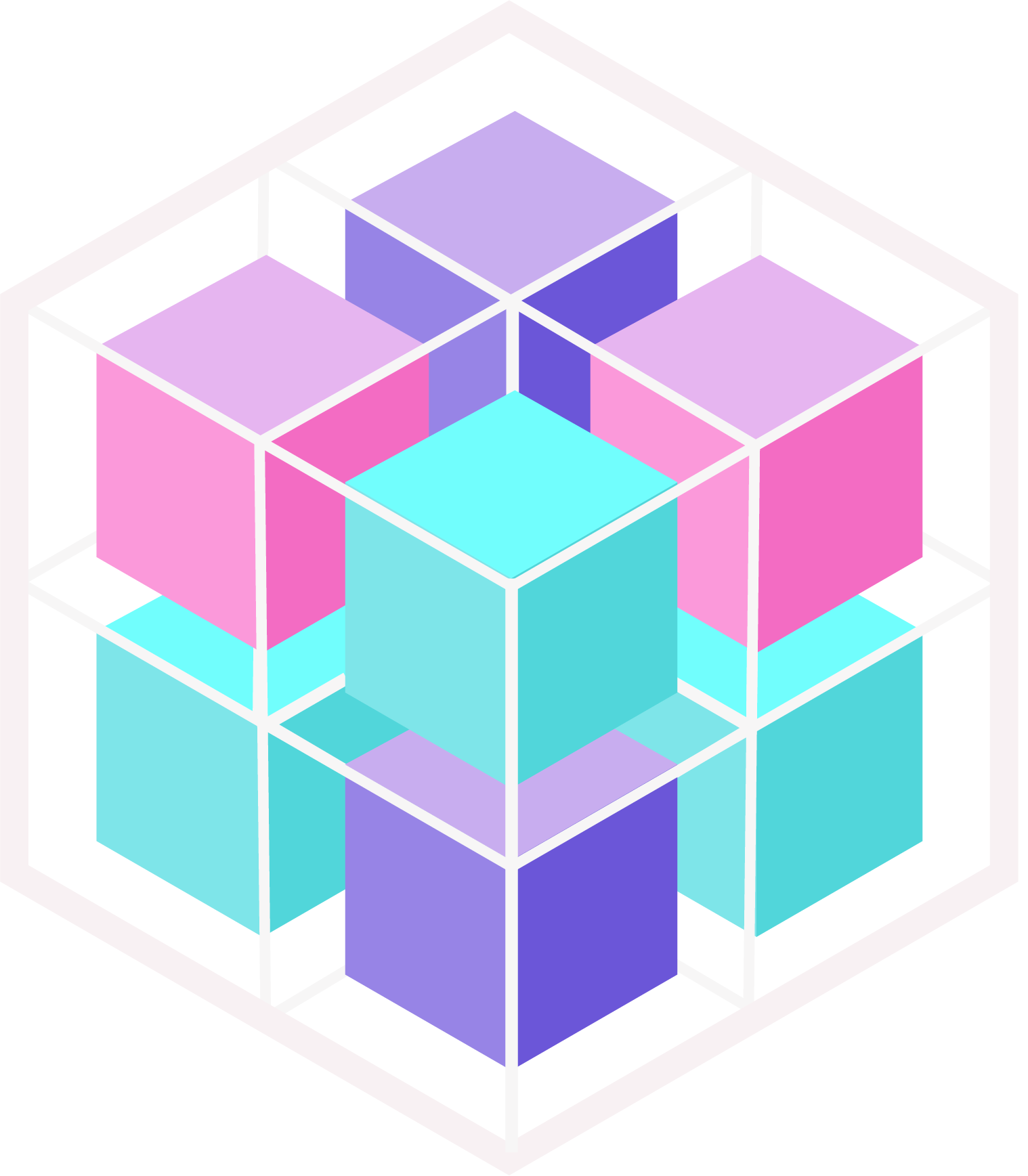}}  \\
\textbf{Federated Learning:} &
  \multicolumn{1}{l|}{} &
  \multicolumn{1}{l|}{} &
  \multicolumn{1}{l|}{} &
  \multicolumn{1}{l|}{} &
  \multicolumn{1}{l|}{} \\
\rowcolor[HTML]{EFEFEF} 
$\quad \bullet $ Use federated models with different datasets &
  \GreenCircle &
  \GreenCircle &
  \GreenCircle &
  \GreenCircle &
  \GreenCircle \\
$\quad \bullet$ Support for other libraries &
  \GreenCircle &
  \RedCircle &
  \GrayCircle &
  \GreenCircle &
  \RedCircle \\
\rowcolor[HTML]{EFEFEF} 
$\quad \bullet$ Sampling environment: IID or non-IID distribution &
  \GreenCircle &
  \RedCircle &
  \GrayCircle &
  \GrayCircle &
  \GrayCircle \\
$\quad \bullet$ Federated aggregation mechanisms &
  \OrangeCircle &
  \OrangeCircle &
  \OrangeCircle &
  \OrangeCircle &
  \OrangeCircle \\
\rowcolor[HTML]{EFEFEF} 
$\quad \bullet$ Federated attack simulator &
  \RedCircle &
  \RedCircle &
  \RedCircle &
  \RedCircle &
  \RedCircle \\
\textbf{Differential Privacy:} &
   &
   &
   &
   &
   \\
\rowcolor[HTML]{EFEFEF} 
$\quad \bullet$ Mechanisms: Exponential, Laplacian, Gaussian &
  \RedCircle &
  \RedCircle &
  \RedCircle &
  \RedCircle &
  \RedCircle \\
$\quad \bullet$ Sensitivity sampler &
  \RedCircle &
  \RedCircle &
  \RedCircle &
  \RedCircle &
  \RedCircle \\
\rowcolor[HTML]{EFEFEF} 
$\quad \bullet$ Subsampling methods to increase privacy &
  \RedCircle &
  \RedCircle &
  \RedCircle &
  \RedCircle &
  \RedCircle \\
$\quad \bullet$ Adaptive Differential Privacy &
  \RedCircle &
  \RedCircle &
  \RedCircle &
  \RedCircle &
  \RedCircle \\
\rowcolor[HTML]{EFEFEF}  
\textbf{Desired properties:} &
  \multicolumn{1}{l|}{} &
  \multicolumn{1}{l|}{} &
  \multicolumn{1}{l|}{} &
  \multicolumn{1}{l|}{} &
  \multicolumn{1}{l|}{} \\
$\quad \bullet$ Documentation \& tutorials &
  \GreenCircle &
  \OrangeCircle &
  \OrangeCircle &
  \OrangeCircle &
  \RedCircle \\
\rowcolor[HTML]{EFEFEF}  
$\quad \bullet$ High level API &
  \RedCircle &
  \GreenCircle &
  \GreenCircle &
  \GrayCircle &
  \GrayCircle \\
$\quad \bullet$ Ability to extend the framework with new properties &
  \GreenCircle &
  \GreenCircle &
  \OrangeCircle&
  \GrayCircle &
  \GrayCircle \\ \hline
\end{tabular}%
}
\caption{ \correcciones{FL and DP features comparison among existing frameworks. \\ \GreenCircle  \Hquad Complete \Hquad \OrangeCircle \Hquad  Partial \Hquad \RedCircle \Hquad  Do not work \Hquad  \GrayCircle \Hquad Unknown}}
\label{tab:comparison_without_sherpa}
\end{table}

\section{\texttt{Sherpa.ai} Federated Learning Framework}
\label{sec:software}

\correcciones{We develop \sherpa{\texttt{Sherpa.ai} FL},\footnote{\url{https://developers.sherpa.ai/privacy-technology/}}\textsuperscript{,}\footnote{\url{https://github.com/sherpaai/Sherpa.ai-Federated-Learning-Framework}} which is an open-research unified FL and DP framework that aims to foster the research and development of AI services at the edges and to preserve data privacy. We describe the hierarchical and modular software architecture of \sherpa{\texttt{Sherpa.ai} FL}, related to the key elements of FL and DP shown in Section \ref{s_soft_arq}. Likewise, we detail the functionalities and the implementation details of \sherpa{\texttt{Sherpa.ai} FL} in Section \ref{ss_soft_funct} and Section \ref{ss_imp_details}.}









\subsection{Software architecture}
\label{s_soft_arq}
The software is structured in several modules that encapsulate the specific functionality of each key element of a FL setting. 
The architecture of these software modules allows the extension of the framework in relation to the progress of the research on FL. Figure \ref{modules} shows the backbone of the software architecture of \sherpa{\texttt{Sherpa.ai} FL}, and we describe each module as what follows:

\begin{figure*}[!b]
  \centering
  \includegraphics[width = 0.65\linewidth]{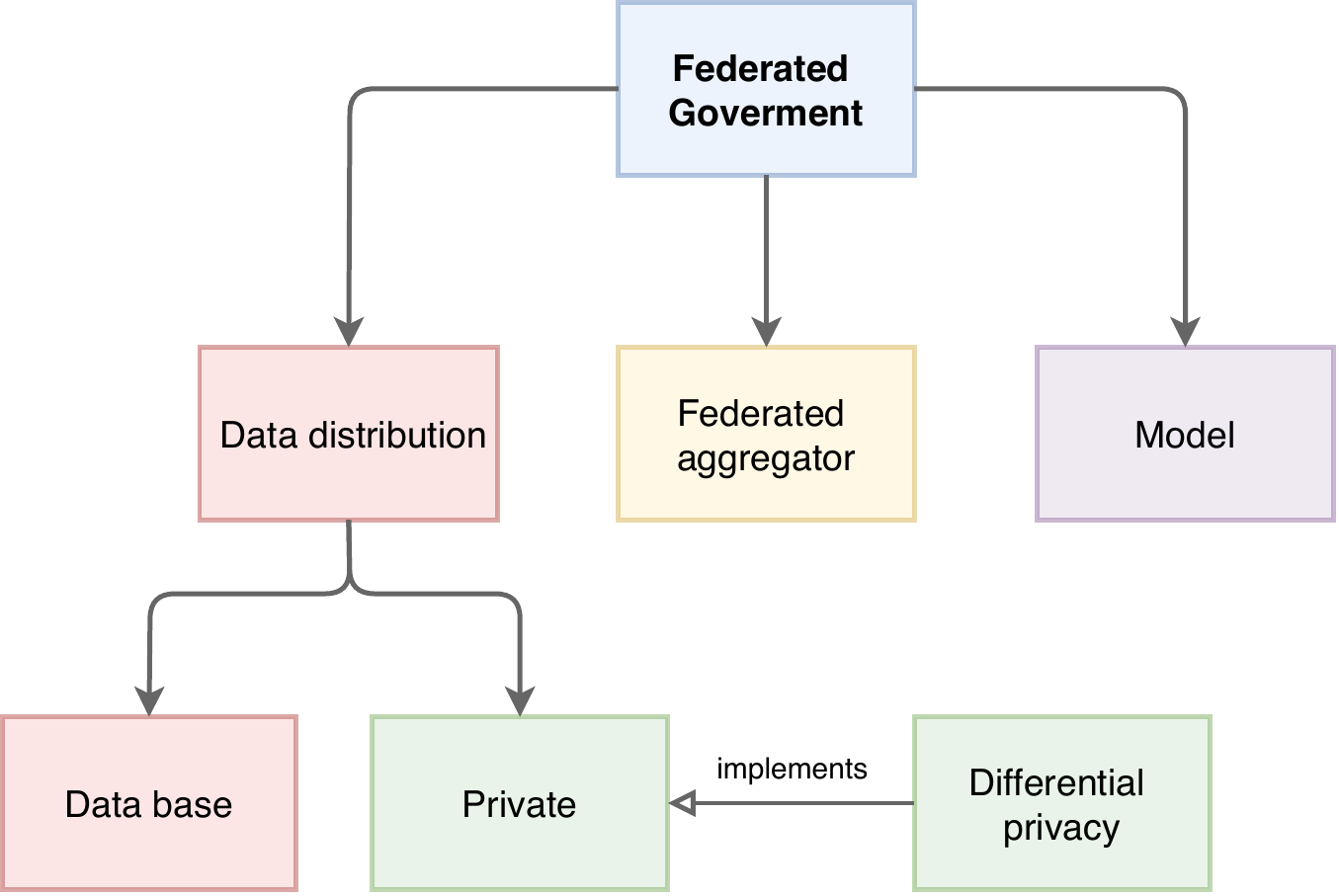}
  \caption{Links between the different modules of \sherpa{\texttt{Sherpa.ai} FL}.}
  \label{modules}
\end{figure*}

\begin{itemize}
    \item \texttt{data\_base}: it is in charge of reading the data according to the chosen database. It is related to the \textit{data} key element.
    \item \texttt{data\_distribution}: it performs the federated distribution of data among the clients involved in the FL process. It is also related to the \textit{data} key element and completes its functionality.
    \item \texttt{private}: it includes several interfaces such as the node interface which represents the \textit{clients} key element and other ones that allow to access and modify the federated data distribution.  
    \item \texttt{learning\_approach}: it represents the whole FL scheme including the federated server model and the communication and coordination among federated server and clients. It encapsulates the \textit{federated server} and the \textit{communication} key elements.
    \item \texttt{federated\_aggregator}: \correcciones{it defines the software structure to develop federated aggregation operators. It is linked to the \textit{federated aggregation operator} key element.}
    \item \texttt{model}: it defines the learning model  using predefined models and their functionalities. This learning model could be any machine learning model that can be aggregated by its representation in parameters. It is related to the \textit{model} key element, as we associate a model object with the clients and the federated server.
    \item \texttt{differential\_privacy}: it preserves DP of the clients by specifying the data access. It is related with the DP key elements, and also with the \textit{data}, the \textit{clients} and the \textit{communication} FL key elements.
\end{itemize}

\subsection{Software functionalities}
\label{ss_soft_funct}
In this section we highlight the main contributions of \sherpa{\texttt{Sherpa.ai} FL}, which are summarised in a wide range of functionalities, namely:


\begin{itemize}
    \item To define and customise a FL simulation with a fixed number of clients using classical data sets.
    \item To define the previous FL simulation using high-level functionalities. 
    \item To train machine learning models among different clients. \correcciones{Currently, \sherpa{\texttt{Sherpa.ai} FL}} offers support for a Keras models (neural networks), and for several models from Scikit-Learn (linear regression, k-means clustering, logistic regression).
    \item To aggregate the information learned from each of the clients into a global model using classical federated aggregation operators such as: FedAvg, weighted FedAvg \cite{bib:mcmahan16} and \correcciones{an aggregation operator for the adaptation of the k-means algorithm to the federated setting \cite{clustering_avg}.}
    \item To apply modifications on federated data such as normalisation or reshaping.
    \item To evaluate the FL approach in comparison with the classical centralised one.
    \item To preserve DP of clients' data and model parameters in the FL context. The platform currently offers support for the fundamental DP mechanisms (Randomized Response, Laplace, Exponential, Gauss), and the composition of DP mechanisms (Basic and Advanced adaptive composition using privacy filters for the maximum privacy loss). Moreover, it is possible to increase privacy by subsampling.  

\end{itemize}

\begin{table}[!t]
\centering
\captionsetup{justification=centering,margin=0.5cm}
\resizebox{\textwidth}{!}{%
\begin{tabular}{|l|c|c|c|c|c|c|}
\rowcolor[HTML]{3b8faa} 
{\color[HTML]{FFFFFF} \textbf{\begin{tabular}[c]{@{}l@{}}FL \& DP features\end{tabular}}} &
  \makecell{\color[HTML]{FFFFFF}
  \includegraphics[width =  0.08 \textwidth]{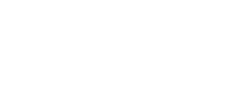}}&
  \makecell{\color[HTML]{FFFFFF}  \textbf{PySyft} \\ \includegraphics[width =  0.02\textwidth]{images/icons/pysyft_logo.png}} &
  \makecell{\color[HTML]{FFFFFF}  \textbf{TensorFlow} \\ \includegraphics[width =  0.03\textwidth]{images/icons/tensorflow-icon-67-32x32.png}} &
  \makecell{\color[HTML]{FFFFFF} \includegraphics[width = 0.05\textwidth]{images/icons/FATE_logo.png}}  &
  \makecell{\color[HTML]{FFFFFF}  \textbf{LEAF} \\ \includegraphics[width = 0.02\textwidth]{images/icons/leaf_icon.png}}  &
  \makecell{\color[HTML]{FFFFFF}  \textbf{PaddleFL} \\ \includegraphics[width = 0.02\textwidth]{images/icons/paddlefl_logo.png}}  \\
\textbf{Federated Learning:} &
   &
  \multicolumn{1}{l|}{} &
  \multicolumn{1}{l|}{} &
  \multicolumn{1}{l|}{} &
  \multicolumn{1}{l|}{} &
  \multicolumn{1}{l|}{} \\
\rowcolor[HTML]{EFEFEF} 
$\quad \bullet $ Use federated models with different datasets &
  \GreenCircle &
  \GreenCircle &
  \GreenCircle &
  \GreenCircle &
  \GreenCircle &
  \GreenCircle \\
$\quad \bullet$ Support for other libraries &
  \OrangeCircle &
  \GreenCircle &
  \RedCircle &
  \GrayCircle &
  \GreenCircle &
  \RedCircle \\
\rowcolor[HTML]{EFEFEF} 
$\quad \bullet$ Sampling environment: IID or non-IID distribution &
  \GreenCircle &
  \GreenCircle &
  \RedCircle &
  \GrayCircle &
  \GrayCircle &
  \GrayCircle \\
$\quad \bullet$ Federated aggregation mechanisms &
  \GreenCircle &
  \OrangeCircle &
  \OrangeCircle &
  \OrangeCircle &
  \OrangeCircle &
  \OrangeCircle \\
\rowcolor[HTML]{EFEFEF} 
$\quad \bullet$ Federated attack simulator &
  \GreenCircle &
  \RedCircle &
  \RedCircle &
  \RedCircle &
  \RedCircle &
  \RedCircle \\
\textbf{Differential Privacy:} &
   &
   &
   &
   &
   &
   \\
\rowcolor[HTML]{EFEFEF} 
$\quad \bullet$ Mechanisms: Exponential, Laplacian, Gaussian &
  \GreenCircle &
  \RedCircle &
  \RedCircle &
  \RedCircle &
  \RedCircle &
  \RedCircle \\
$\quad \bullet$ Sensitivity sampler &
  \GreenCircle &
  \RedCircle &
  \RedCircle &
  \RedCircle &
  \RedCircle &
  \RedCircle \\
\rowcolor[HTML]{EFEFEF} 
$\quad \bullet$ Subsampling methods to increase privacy &
  \GreenCircle &
  \RedCircle &
  \RedCircle &
  \RedCircle &
  \RedCircle &
  \RedCircle \\
$\quad \bullet$ Adaptive Differential Privacy &
  \GreenCircle &
  \RedCircle &
  \RedCircle &
  \RedCircle &
  \RedCircle &
  \RedCircle \\
\rowcolor[HTML]{EFEFEF}  
\textbf{Desired properties:} &
  \multicolumn{1}{l|}{} &
  \multicolumn{1}{l|}{} &
  \multicolumn{1}{l|}{} &
  \multicolumn{1}{l|}{} &
  \multicolumn{1}{l|}{} &
  \multicolumn{1}{l|}{} \\
$\quad \bullet$ Documentation \& tutorials &
  \GreenCircle &
  \GreenCircle &
  \OrangeCircle &
  \OrangeCircle &
  \OrangeCircle &
  \RedCircle \\
\rowcolor[HTML]{EFEFEF}  
$\quad \bullet$ High level API &
  \GreenCircle &
  \RedCircle &
  \GreenCircle &
  \GreenCircle &
  \GrayCircle &
  \GrayCircle \\
$\quad \bullet$ Ability to extend the framework with new properties &
  \GreenCircle &
  \GreenCircle &
  \GreenCircle &
  \OrangeCircle&
  \GrayCircle &
  \GrayCircle \\ \hline
\end{tabular}%
}
\caption{ FL \& DP features comparison between existing frameworks and \sherpa{\texttt{Sherpa.ai} FL}. \\ \GreenCircle  \Hquad Complete \Hquad \OrangeCircle \Hquad  Partial \Hquad \RedCircle \Hquad  Do not work \Hquad  \GrayCircle \Hquad Unknown}
\label{tab:comparison}
\end{table}

In Table \ref{tab:comparison}, we summarise the main contributions of \sherpa{\texttt{Sherpa.ai} FL} in comparison with the key points analysed for each framework in the previous Section.

\correcciones{Thanks to the hierarchical implementation of each module, the aforementioned functionalities can be extended and customised just by adding software classes that inherit from the original software classes.}
For example, the already available machine learning models, DP mechanisms and federated aggregation operators can be modified, or new ones can be created, simply by overwriting the corresponding methods in the classes \texttt{TrainableModel}, \texttt{DataAccessDefinition},  \texttt{FederatedAggregator}, respectively.

\subsection{Implementation details}
\label{ss_imp_details}

\sherpa{\texttt{Sherpa.ai} FL} has been developed by DaSCI\footnote{\url{https://dasci.es/}} \correcciones{Institute} and Sherpa.ai.\footnote{\url{https://sherpa.ai/}} We developed the software using Python language for the whole architecture. Furthermore, Keras\footnote{\url{https://keras.io/}}, TensorFlow\footnote{\url{https://www.tensorflow.org/}} and scikit-kearn\footnote{\url{https://scikit-learn.org/stable/}} APIs are employed for the machine learning part which ensures efficiency and compatibility. 

It can also be run on computing devices such as CPUs or GPUs. In order to use GPUs, the adequate versions of TensorFlow and CUDA must be installed. For detailed installation instructions, please see the installation guide.\footnote{\url{https://github.com/sherpaai/Sherpa.ai-Federated-Learning-Framework/blob/master/install.md}}\textsuperscript{,}\footnote{The implementation details described in this paper corresponds to the release 0.1.0 of \sherpa{\texttt{Sherpa.ai} FL}.}


\correcciones{The framework is licensed under the Apache License 2.0,\footnote{\url{https://www.apache.org/licenses/LICENSE-2.0}} a permissive license whose main conditions require preservation of copyright and license notices.}

\section{\correcciones{Machine learning matches federated learning.  Methodological guidelines for preserving data privacy}}
\label{s_ml_fl}

\correcciones{FL is a paradigm of machine learning, but its particularities force to adapt the machine learning settings to FL. For this reason, \sherpa{\texttt{Sherpa.ai} FL} functionalities outlined in  Section \ref{sec:software} stem from the need to develop machine learning algorithms specialised for Federated Artificial Intelligence that protect clients' data privacy. However, the adaptation is not only focused on the algorithms, but also on the workflow of machine learning.

In this section, we first discuss the key aspects of distributed computing for machine learning in the federated setting. Following, Section \ref{sec:fml_paradigm} defines a rather specific paradigm for adapting machine learning models to the federated setting,  followed by some remarkable exceptions to the defined adaptation in Section \ref{sec_fml_non_paradigm}. Finally, we also define the adaptation of the machine learning workflow to FL in Section \ref{sec:guidelines} as methodological guidelines for preserving data privacy with FL using \sherpa{\texttt{Sherpa.ai} FL}.


}

\subsection{Key aspects of distributed computing for Federated Machine Learning}\label{sec:fml_key_aspects}

The recent introduction of FL   \citep{bib:konecny16federatedoptimizacion, bib:mcmahan16communicationefficient, bib:konecny16federatedlearning}
responds to the need for novel distributed machine learning algorithms in a setting that clashes with several assumptions of conventional parallel machine learning in a data centre. The differences are substantially originated by the unreliable and poor network connection of the clients, since clients are typically mobile phones. Thus reducing the number of \correcciones{rounds of learning} is \textit{essential} as communication constrains are more severe. 
Additionally, the data is unevenly scattered across $K$ clients and must be considered as non-IID, that is, the data accessible locally is not in any way representative of the overall trend.\footnote{\sherpa{\texttt{Sherpa.ai} FL} allows for both IID and non-IID client data.}
The data is in general sparse, where the features of interest take place on a reduced number of clients or data points.\footnote{\sherpa{\texttt{Sherpa.ai} FL} allows for \textit{weighted} aggregation, emphasising the contribution of most significant clients to the global model.}
Ultimately, the number of total clients greatly exceeds the number of training points available locally on each client ($K \gg n/K$).

In the federated machine learning setting, the training is decoupled from the access to the raw data. In fact, the raw data never leaves users' mobile devices and a high-accuracy model is produced in a central server by aggregating locally computed updates. At each FL round, an update vector $\theta \in \mathbb{R}^d$ is sent from each client to the central server to improve the global model, with $d$ the  parameters' dimension of the computed model.
It is worth noting that the magnitude of the update $\theta$ is thus independent from the amount of raw data available on the local client (\correcciones{\em{e.g.},} $\theta$ might be a gradient vector). 
One of the advantages of this approach \correcciones{is} the considerable bandwidth and time saved in data communication.

Another motivation for the FL setting (but that also constitutes one of its intrinsic advantages) is the concern for privacy and security.
By not transferring any raw data to the central server, the attack surface reduces to only the single client, instead of both client and server. On the other hand, the update $\theta$ sent by the client might still reveal some of its private information, however the latter will be almost always dramatically reduced with respect to the raw training data.   
Besides, after improving the current model by the update $\theta$, this can (and should) be deleted.

Additional privacy can be provided by randomised algorithms providing DP \citep{TCS-042}, as detailed in Sections \ref{sec:dp_definition} and \ref{sec:dp_key_elements}.
In particular, the centralised algorithm could be equipped with a DP layer allowing the release of the global model without compromising the privacy of the individual clients who contributed to its training (\st{see} \correcciones{\em{e.g.},} \citet{Abadi2016}). On the other hand, in the case of a malicious or compromised server, or in the case of potential eavesdropping, DP can be applied on the local clients\footnote{\sherpa{\texttt{Sherpa.ai} FL} allows to apply sophisticated and customised DP mechanisms on the model's parameters, as well as on client's raw data (see Section \ref{sec:dp_key_elements}).} for protecting their privacy \citep{Duchi2012, Geyer2017, Jiang2019}.

\subsection{The Machine Learning paradigm in a federated setting}
\label{sec:fml_paradigm}
In the following, we describe the \textit{federated machine learning paradigm} by recognising some relevant attributes that ease the natural adaptation of a ML model in the federated setting. 

Primarily, we observe that a great number of machine learning methods resemble the minimisation of an objective function with finite-sum as in Equation \ref{eq_fl_objective_function}.
The aforementioned problem structure encompasses both linear and logistic regressions, support vector machines, and also more elaborated techniques such as conditional random fields and neural networks \citep{bib:konecny16federatedoptimizacion}. 
Indeed, in neural networks predictions are made through a non-convex function, yet the resulting objective function can still be expressed as $F_i(\theta)$ and the gradients can be efficiently obtained by backpropagation, thus resembling Equation \ref{eq_fl_objective_function}.

A variety of algorithms have been proposed to solve the minimisation problem in Equation \ref{eq_fl_objective_function} in the federated setting, where, as mentioned earlier, the primary constrain is the communication efficiency for reducing the number of FL rounds in the aggregation of local models. In this context, another characteristic trait of the Federated ML paradigm is constituted by the intrinsic compatibility with baseline aggregation operators (e.g. Federated Averaging), and where no ad-hoc adaptation is required.

Ultimately, several of these FL algorithms have been supplied with DP  \citep{Abadi2016, Geyer2017, Jiang2019}. We thus identify a rather important aspect of the Federated machine learning paradigm as being prone to straightforward application of the common building components of DP.  In addition, the latter feature eases the task of estimating the privacy loss in the FL rounds by the application of composition theorems for DP.\footnote{\sherpa{\texttt{Sherpa.ai} FL} offers support for both common building components for DP, as well as for its basic and advanced composition theorems using privacy filters (see Section \ref{sec:dp_key_elements}).}

To summarise, a machine learning method is prone to adaptation in the federated setting if it adheres to the principles of the federated machine learning paradigm described above, namely: \begin{enumerate*}[label={(\arabic*)}] \item the  problem structure resembling the minimisation of an objective function as in  Equation \ref{eq_fl_objective_function}, \item the attribute of easy aggregation of local models' parameters, and \item the direct applicability of DP techniques for additional privacy.\end{enumerate*} Among such machine learning models we cite neural networks \citep{bib:bonawitz19}, linear \citep{bib:Gascn2016, bib:Gascn2017} and logistic \citep{bib:Hardy2017} regressions.\footnote{See also notebooks on deep learning, linear and logistic regressions available in \sherpa{Sherpa.ai FL} at \url{https://github.com/sherpaai/Sherpa.ai-Federated-Learning-Framework/tree/master/notebooks}}

\subsection{Models deviating from the federated machine learning paradigm} \label{sec_fml_non_paradigm}
It is worth mentioning specific machine learning models whose structure only partially fits in the federated machine learning paradigm described above. Although the problem structure can still be represented by a minimisation of an objective function as in Equation \ref{eq_fl_objective_function}, their adaptation to a federated setting requires additional and ad-hoc procedures. 

One example is found in the \textit{k-means clustering} algorithm for unsupervised learning \citep{Lloyd1982, James2013}, where  the non-IID nature of the data distribution is seen as a major obstacle in a federated setting. 
Namely, the direct application of average aggregation is unfeasible due to potentially different  number  and ordering of local clusters, and more advanced algorithms need to be employed. A workable solution is to fix the number of local clusters, and apply an additional k-means clustering in the average aggregation.\footnote{See notebook on k-means clustering in \sherpa{Sherpa.ai FL} at \url{https://github.com/sherpaai/Sherpa.ai-Federated-Learning-Framework/tree/master/notebooks}} Alternatively, one might try grouping the clients' population sharing jointly trainable data distributions, as proposed by \citet{Sattler2019} in the context of deep neural networks. An additional complication is constituted by the preservation of clients' privacy. 
For instance, the baseline DP building components necessitate some adjustments in order to  be applied. Although not in a FL context, in \citet{Zhang2018} the authors adjust the Laplace noise added to each centroid based on the contour coefficients.

Another notable example is represented by the federated version of matrix factorization-based ranking algorithms for recommendation systems \citep{Funk2006, Ammaduddin2019}. The peculiar architecture of this algorithm involves the communication of only a portion of the update vector for improving the model, thus the round of learning implies additional communications between the central server and the clients. Moreover, the multiple communication iterations necessitates further caution with privacy loss in the DP context. A viable approach is to implement a two-stage randomised response mechanism on the local data, and to allow the clients to specify their privacy level \citep{Jiang2019}.


\subsection{\correcciones{Methodological guidelines for preserving data privacy with federated learning in \sherpa{\texttt{Sherpa.ai} FL}}} \label{sec:guidelines}

\correcciones{The experimental settings of FL and machine learning are very similar because FL is a machine learning paradigm. Nonetheless, the particularities of FL force to revise the machine learning workflow, and adapt it to the FL definition. In this section, we define the workflow of FL based on the machine learning workflow, and we present it as methodological guidelines. \sherpa{\texttt{Sherpa.ai} FL} comply with these methodological guidelines assuring the following of good practises in the development of AI services at the edges that preserve data privacy.}


\correcciones{We distinguish two scenarios in FL}: \begin{enumerate*}[label={(\arabic*)}]\item a real one, where we do not actually know the underlying distribution of the data and \item a simulation of a FL scenario, where it is possible to emulate a federated data distribution in order to analyse its use case. \end{enumerate*} The guidelines are focused on the real FL scenario although we remark the particularities of a simulated FL experiment. 

\begin{figure*}[h!]
  \centering
  \includegraphics[width = \linewidth]{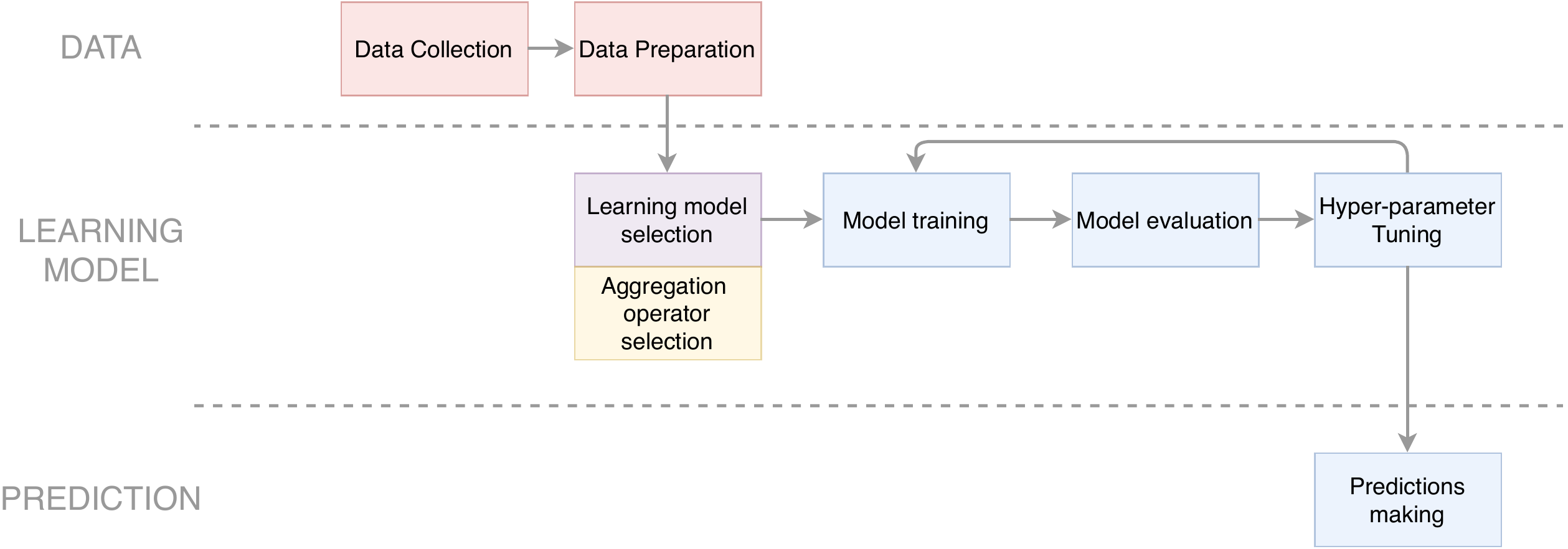}
  \caption{\correcciones{Flow chart} of a FL experiment.}
  \label{steps}
\end{figure*}

Moreover, we assume that the problem is properly formulated, that is, the data features and the target variable are previously defined and agreed upon by the clients. Based on this hypothesis, we show the scheme of the workflow of a FL experiment in Figure \ref{steps} and detail it \correcciones{in the following sections}.

\subsubsection{Data collection}

In a real FL scenario, the data naturally belongs to the clients. Therefore, the data collection takes place locally at each client, resulting in a distributed approach from the outset. In a strict FL scenario, the server has no knowledge at all of the data. However, there is the possibility for the server to gain minor prior knowledge of the problem if a global validation or test dataset is used. Here, we assume that the server does not have any information, which is the most restrictive and common situation.

\textit{Remark:} When simulating a FL scenario in scientific research, data collection is reduced to accessing a database. The distribution of the data among clients is simulated in the data preparation step.

\subsubsection{Data preparation}

Data preparation involves two tasks: \begin{enumerate*}[label={(\arabic*)}]\item Data partition, where we split data in training, evaluation and test sets, and \item data preprocessing, where we transform the training data in order to improve its quality.\end{enumerate*} 

\paragraph{\textbf{Data partition}}


The process of splitting data into training, evaluation and test datasets in FL is similar to centralised machine learning process with the difference of replicating the process for the data stored on each client. That is, each client dataset is split into training, evaluation and test sets.

\textit{Remark}: When it comes to a FL scenario for scientific research, it is feasible to have global evaluation and test datasets by extracting them before assigning the rest of the data to the clients as local training datasets. Moreover, in a simulation it could be a good practise to use both global and local evaluation and test datasets combining both methodologies.

\paragraph{\textbf{Data preprocessing}}

Preprocessing is the most tricky task in FL due to the distributed and private character of the data. The challenge is to consistently preprocess distributed datasets in several clients without any clue about the underlying data distribution.

The process of adapting centralised preprocessing techniques to federated data is time-consuming. For the techniques based on statistics of data distributions (e.g. normalisation) it is necessary to use robust aggregation of the statistics, which is a challenge in some situations. Algorithms based on intervals (e.g. discretisation) require a global interval that includes all the possible values. Moreover, there are complicated methods of robust adaptation such as feature selection \cite{feature_selection}. Because of these intricacies, it is advisable to rely on preprocessing techniques adapted to distributed scenarios.

Regarding distributed data preprocessing, we might take inspiration from different distributed preprocessing techniques that have already been developed \cite{distributed_datamining}. However, most of these methods need to be adapted in order to respect data privacy. A distributed model that suits privacy restrictions is MapReduce \cite{mapreduce}. Therefore, big data preprocessing techniques \cite{bigdata} that are interactively applicable can be adapted to a FL scenario in compliance with data constraints.

\textit{Remark}: When simulating FL, it is possible to use centralised preprocessing methods before splitting the data between the clients. It is not a recommended practice, but a useful trade-off in terms of experimentation.

\subsubsection{Model selection}

This step implies, besides the choice of the learning model as in any centralised approach, the choice of the parameter aggregation mechanism used in the server.

\paragraph{\textbf{Choosing the learning model}}
This task consists of choosing the learning model structure stored both in the server and the clients. Clearly, the model has to correspond to the type of problem being addressed. The only restriction is that the learning model has to be representable using parameters in order to get the server learning model by aggregating local parameters. The canonical example of a learning model that can be represented using parameters is deep learning, but this is not the only one. 

\textit{Remark}: When we simulate a FL scenario, server learning model can be initialised using global previous information. However, in a strict FL scenario it is initialised with the first aggregation of local parameters.

\paragraph{\textbf{Choosing the aggregation operator}}

We also need to choose the aggregation operator used for client parameters aggregation at this point. There are different types of aggregation operators: \begin{enumerate*}[label={(\arabic*)}]\item operators which aggregate every client parameters (such as FedAvg), \item operators which select the clients that take part in the aggregation (e.g. based on the performance) and \item asynchronous aggregation operators (such as CO-OP). \end{enumerate*} 

\subsubsection{Model training}

The iterative FL training process is divided into rounds of learning, and each round consists of: \begin{enumerate}
\item Training the local models on their local training dataset,

\item sharing of the local parameters to the server, 

\item aggregation of local models' parameters  on the server using the aggregation operator and

\item updating the local models with the aggregated global model.
\end{enumerate}

\subsubsection{Model evaluation}

The evaluation of a FL model consists of assessing the aggregated model after assigning it to each client using the local evaluation datasets. After that, each client shares the performance with the server, which combines the local performances resulting in global evaluation metrics. Since the amount of data per client can be variable, we recommend using absolute metrics on clients (e.g. confusion matrix) and combine them on the server to get the remaining evaluation metrics.

\textit{Remark 1}: When simulating FL, we can use a global evaluation dataset in order to evaluate the performance of the aggregated model. Moreover, we can use cross-validation methodologies to evaluate the model's performance by partitioning all the folds at the beginning and replicating the whole workflow for each of the fold combinations.

\textit{Remark 2}: Although it is not the main purpose of FL, it might be worthwhile to evaluate the local models prior to the aggregation for measuring the customisation of the local model to each client.

\subsubsection{Hyper-parameter tuning}

We base the tuning of the hyper-parameters of the learning models on the metrics obtained in the previous step, and modify certain learning model parameters in order to improve the performance on the evaluation datasets. 

\textit{Remark}: According to the previously mentioned customisation, although it is not the objective of the FL, we could tune each of the local models independently according to the local evaluation performance before the aggregation in order to improve customisation.

\subsubsection{Predictions making}

\correcciones{The last step in the machine learning workflow after the training of the learning model, and by extension in the corresponding FL one,  is to predict the label of unknown examples. Those predictions are done with test sets of each client.}


\textit{Remark}: When simulating FL, we can use global test dataset for prediction. Moreover, we can test local learning models prior to aggregation using instances of other clients (unknown targets) in order to measure the capability of generalisation of local models.

\section{Illustrative cases of study}\label{sec:illustrative-example}
\correcciones{One of the main characteristics of \sherpa{\texttt{Sherpa.ai} FL} is its development upon the methodological guidelines for FL detailed in Section \ref{sec:guidelines}. In this section, we show how to follow these methodological guidelines with \sherpa{\texttt{Sherpa.ai} FL} through two experimental use cases, namely:}

\begin{enumerate}
\item \correcciones{Classification with FL (see Section \ref{sec:emnist})}: showing how to create each of the key-elements of a FL experiment and combine them using our framework. To end with this example, we also compare the FL approach with a centralised one.
\item \correcciones{Regression with FL and DP} (see Section \ref{sec:linear_regressionDP}):  we compare the centralised with the FL approach with DP. Moreover, we demonstrate how to limit the privacy loss through the Privacy Filters implemented in \sherpa{\texttt{Sherpa.ai} FL}. \end{enumerate}

For more illustrative examples of the framework use, please see the notebook examples.\footnote{\url{https://github.com/sherpaai/Sherpa.ai-Federated-Learning-Framework/tree/master/notebooks}}


\subsection{Classification with FL}\label{sec:emnist}


In this section we provide a simple example of how to \correcciones{develop a classification experiment in a FL setting with} \sherpa{\texttt{Sherpa.ai} FL}. We use a popular dataset to start the experimentation in a federated environment, and we finish the example with a comparison between federated and centralised approaches. 

\begin{figure}[htp!]
  \centering
  \includegraphics[width=\linewidth]{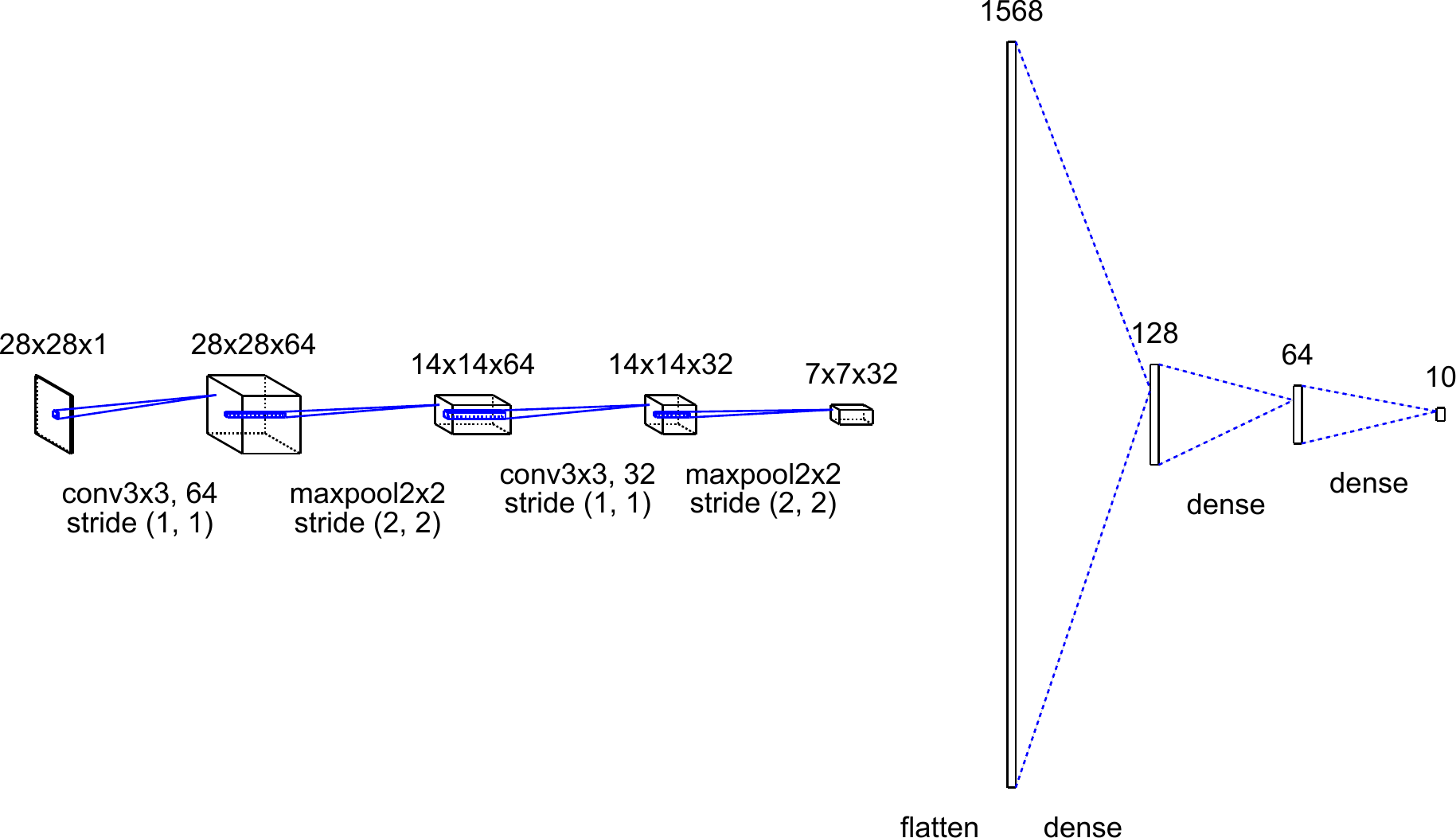}
  \caption{CNN-based neural network used as learning model in the illustrative example.}
  \label{fig:cnn}
\end{figure}

\subsubsection{Case of study}

In order to show the functionality of the software, we implement a simple and instructive case of study. \paco{W}e use the EMNIST\footnote{\url{https://www.nist.gov/itl/products-and-services/emnist-dataset}} Digits dataset \cite{cohen2017}. It consists of an extended version of the classic MNIST dataset which includes writings of several authors with different features. This fact provides the non-IID character to the data which is useful for the simulation of federated environments. \eugenio{ Table \ref{tab:distribution} shows the size of the dataset.}

\begin{table}[hbt]
\centering
\begin{tabular}{@{}rrr@{}} \toprule
\textbf{Train set} & \textbf{Test set} & \textbf{Total} \\\midrule
240\,000 & 40\,000 & 280\,000 \\
 \bottomrule
\end{tabular}
\caption{Distribution of EMNIST Digits dataset.}\label{tab:distribution} 
\end{table}

For the simulation of the FL scenario we use 5 clients among which the instances of the dataset are distributed following a non-IID distribution. We use as learning model a simple CNN \paco{(Convolutional Neural Networks)} based neural network represented in Figure \ref{fig:cnn}, and as federated aggregation operator the widely used operator FedAvg. The code of the illustrative example is detailed in the following section.

\subsubsection{Description of the code}

We start the simulation \correcciones{with the first step of the methodological guidelines, \textit{i.e.} the preprocessing of the \textit{data collection}. Accordingly, we begin with} loading the dataset. \texttt{Sherpa.FL} provides some functions to load the EMNIST Digits dataset.

    \begin{tcolorbox}[breakable, size=fbox, boxrule=1pt, pad at break*=1mm,colback=cellbackground, colframe=cellborder]
\prompt{In}{incolor}{1}{\boxspacing}
\begin{Verbatim}[commandchars=\\\{\}]
\PY{k+kn}{import} \PY{n+nn}{matplotlib}\PY{n+nn}{.}\PY{n+nn}{pyplot} \PY{k}{as} \PY{n+nn}{plt}
\PY{k+kn}{import} \PY{n+nn}{shfl}
\PY{k+kn}{from} \PY{n+nn}{shfl}\PY{n+nn}{.}\PY{n+nn}{private}\PY{n+nn}{.}\PY{n+nn}{reproducibility} \PY{k+kn}{import} \PY{n}{Reproducibility}

\PY{c+c1}{\PYZsh{} Comment to turn off reproducibility:}
\PY{n}{Reproducibility}\PY{p}{(}\PY{l+m+mi}{1234}\PY{p}{)}

\PY{n}{database} \PY{o}{=} \PY{n}{shfl}\PY{o}{.}\PY{n}{data\PYZus{}base}\PY{o}{.}\PY{n}{Emnist}\PY{p}{(}\PY{p}{)}
\PY{n}{train\PYZus{}data}\PY{p}{,} \PY{n}{train\PYZus{}labels}\PY{p}{,} \PY{n}{test\PYZus{}data}\PY{p}{,} \PY{n}{test\PYZus{}labels} \PY{o}{=} \PY{n}{database}\PY{o}{.}\PY{n}{load\PYZus{}data}\PY{p}{(}\PY{p}{)}
\end{Verbatim}
\end{tcolorbox}

We can inspect some properties of the loaded data\eugenio{, for instance the size or the dimension of the data}.

    \begin{tcolorbox}[breakable, size=fbox, boxrule=1pt, pad at break*=1mm,colback=cellbackground, colframe=cellborder]
\prompt{In}{incolor}{2}{\boxspacing}
\begin{Verbatim}[commandchars=\\\{\}]
\PY{n+nb}{print}\PY{p}{(}\PY{n+nb}{len}\PY{p}{(}\PY{n}{train\PYZus{}data}\PY{p}{)}\PY{p}{)}
\PY{n+nb}{print}\PY{p}{(}\PY{n+nb}{len}\PY{p}{(}\PY{n}{test\PYZus{}data}\PY{p}{)}\PY{p}{)}
\PY{n+nb}{print}\PY{p}{(}\PY{n+nb}{type}\PY{p}{(}\PY{n}{train\PYZus{}data}\PY{p}{[}\PY{l+m+mi}{0}\PY{p}{]}\PY{p}{)}\PY{p}{)}
\PY{n}{train\PYZus{}data}\PY{p}{[}\PY{l+m+mi}{0}\PY{p}{]}\PY{o}{.}\PY{n}{shape}
\end{Verbatim}
\end{tcolorbox}

    \begin{Verbatim}[commandchars=\\\{\}]
240000
40000
<class 'numpy.ndarray'>
    \end{Verbatim}

            \begin{tcolorbox}[breakable, size=fbox, boxrule=.5pt, pad at break*=1mm, opacityfill=0]
\prompt{Out}{outcolor}{2}{\boxspacing}
\begin{Verbatim}[commandchars=\\\{\}]
(28, 28)
\end{Verbatim}
\end{tcolorbox}

As we see, our dataset is composed by a set of matrix of 28
by 28. Before starting with the federated scenario, we can take a look
to a sample of the training data.

    \begin{tcolorbox}[breakable, size=fbox, boxrule=1pt, pad at break*=1mm,colback=cellbackground, colframe=cellborder]
\prompt{In}{incolor}{3}{\boxspacing}
\begin{Verbatim}[commandchars=\\\{\}]
\PY{n}{plt}\PY{o}{.}\PY{n}{imshow}\PY{p}{(}\PY{n}{train\PYZus{}data}\PY{p}{[}\PY{l+m+mi}{0}\PY{p}{]}\PY{p}{)}
\end{Verbatim}
\end{tcolorbox}

            \begin{tcolorbox}[breakable, size=fbox, boxrule=.5pt, pad at break*=1mm, opacityfill=0]
\prompt{Out}{outcolor}{3}{\boxspacing}
\begin{Verbatim}[commandchars=\\\{\}]
<matplotlib.image.AxesImage at 0x105ea3450>
\end{Verbatim}
\end{tcolorbox}
        
    \begin{center}
    \adjustimage{max size={0.4\linewidth}{0.4\paperheight}}{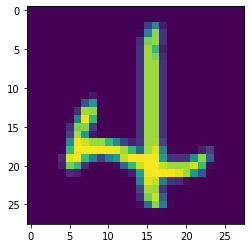}
    \end{center}
    { \hspace*{\fill} \\}
    
Now, we simulate a FL scenario with a set of 5 client nodes containing private data, and a central server which is responsible to coordinate the different clients. First of all, we simulate the data contained in every client with a IID distribution of the data.

    \begin{tcolorbox}[breakable, size=fbox, boxrule=1pt, pad at break*=1mm,colback=cellbackground, colframe=cellborder]
\prompt{In}{incolor}{4}{\boxspacing}
\begin{Verbatim}[commandchars=\\\{\}]
\PY{n}{iid\PYZus{}distribution} \PY{o}{=} \PY{n}{shfl}\PY{o}{.}\PY{n}{data\PYZus{}distribution}\PY{o}{.}\PY{n}{IidDataDistribution}\PY{p}{(}\PY{n}{database}\PY{p}{)}
\PY{n}{federated\PYZus{}data}\PY{p}{,} \PY{n}{test\PYZus{}data}\PY{p}{,} \PY{n}{test\PYZus{}labels} \PY{o}{=} \PY{n}{iid\PYZus{}distribution}\PY{o}{.}\PY{n}{get\PYZus{}federated\PYZus{}data}\PY{p}{(}\PY{n}{num\PYZus{}nodes}\PY{o}{=}\PY{l+m+mi}{5}\PY{p}{,} \PY{n}{percent}\PY{o}{=}\PY{l+m+mi}{50}\PY{p}{)}
\end{Verbatim}
\end{tcolorbox}

As a result, we have created federated data from the EMNIST dataset with 5 nodes and using every available data. Hence, the \textit{data collection} process have finished. This data is a set of data nodes containing private data. 
    \begin{tcolorbox}[breakable, size=fbox, boxrule=1pt, pad at break*=1mm,colback=cellbackground, colframe=cellborder]
\prompt{In}{incolor}{5}{\boxspacing}
\begin{Verbatim}[commandchars=\\\{\}]
\PY{n+nb}{print}\PY{p}{(}\PY{n+nb}{type}\PY{p}{(}\PY{n}{federated\PYZus{}data}\PY{p}{)}\PY{p}{)}
\PY{n+nb}{print}\PY{p}{(}\PY{n}{federated\PYZus{}data}\PY{o}{.}\PY{n}{num\PYZus{}nodes}\PY{p}{(}\PY{p}{)}\PY{p}{)}
\PY{n}{federated\PYZus{}data}\PY{p}{[}\PY{l+m+mi}{0}\PY{p}{]}\PY{o}{.}\PY{n}{private\PYZus{}data}
\end{Verbatim}
\end{tcolorbox}

    \begin{Verbatim}[commandchars=\\\{\}]
<class 'shfl.private.federated\_operation.FederatedData'>
5
Node private data, you can see the data for debug purposes but the data remains
in the node
<class 'dict'>
\{'112883278416': <shfl.private.data.LabeledData object at 0x1a486393d0>\}
    \end{Verbatim}
    
As we can see, private data in a node is not accesible directly but the
framework provides mechanisms to use this data in a machine learning
model.

Once data is prepared, the next step is the definition of the neural network architecture (\textit{model selection}) used along the learning process. The framework provides a class to adapt a Keras (or Tensorflow) model to the framework, so you only have to create a function that will act as model builder.

    \begin{tcolorbox}[breakable, size=fbox, boxrule=1pt, pad at break*=1mm,colback=cellbackground, colframe=cellborder]
\prompt{In}{incolor}{6}{\boxspacing}
\begin{Verbatim}[commandchars=\\\{\}]
\PY{k+kn}{import} \PY{n+nn}{tensorflow} \PY{k}{as} \PY{n+nn}{tf}

\PY{k}{def} \PY{n+nf}{model\PYZus{}builder}\PY{p}{(}\PY{p}{)}\PY{p}{:}
    \PY{n}{model} \PY{o}{=} \PY{n}{tf}\PY{o}{.}\PY{n}{keras}\PY{o}{.}\PY{n}{models}\PY{o}{.}\PY{n}{Sequential}\PY{p}{(}\PY{p}{)}
    \PY{n}{model}\PY{o}{.}\PY{n}{add}\PY{p}{(}\PY{n}{tf}\PY{o}{.}\PY{n}{keras}\PY{o}{.}\PY{n}{layers}\PY{o}{.}\PY{n}{Conv2D}\PY{p}{(}\PY{l+m+mi}{32}\PY{p}{,} \PY{n}{kernel\PYZus{}size}\PY{o}{=}\PY{p}{(}\PY{l+m+mi}{3}\PY{p}{,} \PY{l+m+mi}{3}\PY{p}{)}\PY{p}{,} \PY{n}{padding}\PY{o}{=}\PY{l+s+s1}{\PYZsq{}}\PY{l+s+s1}{same}\PY{l+s+s1}{\PYZsq{}}\PY{p}{,} \PY{n}{activation}\PY{o}{=}\PY{l+s+s1}{\PYZsq{}}\PY{l+s+s1}{relu}\PY{l+s+s1}{\PYZsq{}}\PY{p}{,} \PY{n}{strides}\PY{o}{=}\PY{l+m+mi}{1}\PY{p}{,} \PY{n}{input\PYZus{}shape}\PY{o}{=}\PY{p}{(}\PY{l+m+mi}{28}\PY{p}{,} \PY{l+m+mi}{28}\PY{p}{,} \PY{l+m+mi}{1}\PY{p}{)}\PY{p}{)}\PY{p}{)}
    \PY{n}{model}\PY{o}{.}\PY{n}{add}\PY{p}{(}\PY{n}{tf}\PY{o}{.}\PY{n}{keras}\PY{o}{.}\PY{n}{layers}\PY{o}{.}\PY{n}{MaxPooling2D}\PY{p}{(}\PY{n}{pool\PYZus{}size}\PY{o}{=}\PY{l+m+mi}{2}\PY{p}{,} \PY{n}{strides}\PY{o}{=}\PY{l+m+mi}{2}\PY{p}{,} \PY{n}{padding}\PY{o}{=}\PY{l+s+s1}{\PYZsq{}}\PY{l+s+s1}{valid}\PY{l+s+s1}{\PYZsq{}}\PY{p}{)}\PY{p}{)}
    \PY{n}{model}\PY{o}{.}\PY{n}{add}\PY{p}{(}\PY{n}{tf}\PY{o}{.}\PY{n}{keras}\PY{o}{.}\PY{n}{layers}\PY{o}{.}\PY{n}{Dropout}\PY{p}{(}\PY{l+m+mf}{0.4}\PY{p}{)}\PY{p}{)}
    \PY{n}{model}\PY{o}{.}\PY{n}{add}\PY{p}{(}\PY{n}{tf}\PY{o}{.}\PY{n}{keras}\PY{o}{.}\PY{n}{layers}\PY{o}{.}\PY{n}{Conv2D}\PY{p}{(}\PY{l+m+mi}{32}\PY{p}{,} \PY{n}{kernel\PYZus{}size}\PY{o}{=}\PY{p}{(}\PY{l+m+mi}{3}\PY{p}{,} \PY{l+m+mi}{3}\PY{p}{)}\PY{p}{,} \PY{n}{padding}\PY{o}{=}\PY{l+s+s1}{\PYZsq{}}\PY{l+s+s1}{same}\PY{l+s+s1}{\PYZsq{}}\PY{p}{,} \PY{n}{activation}\PY{o}{=}\PY{l+s+s1}{\PYZsq{}}\PY{l+s+s1}{relu}\PY{l+s+s1}{\PYZsq{}}\PY{p}{,} \PY{n}{strides}\PY{o}{=}\PY{l+m+mi}{1}\PY{p}{)}\PY{p}{)}
    \PY{n}{model}\PY{o}{.}\PY{n}{add}\PY{p}{(}\PY{n}{tf}\PY{o}{.}\PY{n}{keras}\PY{o}{.}\PY{n}{layers}\PY{o}{.}\PY{n}{MaxPooling2D}\PY{p}{(}\PY{n}{pool\PYZus{}size}\PY{o}{=}\PY{l+m+mi}{2}\PY{p}{,} \PY{n}{strides}\PY{o}{=}\PY{l+m+mi}{2}\PY{p}{,} \PY{n}{padding}\PY{o}{=}\PY{l+s+s1}{\PYZsq{}}\PY{l+s+s1}{valid}\PY{l+s+s1}{\PYZsq{}}\PY{p}{)}\PY{p}{)}
    \PY{n}{model}\PY{o}{.}\PY{n}{add}\PY{p}{(}\PY{n}{tf}\PY{o}{.}\PY{n}{keras}\PY{o}{.}\PY{n}{layers}\PY{o}{.}\PY{n}{Dropout}\PY{p}{(}\PY{l+m+mf}{0.3}\PY{p}{)}\PY{p}{)}
    \PY{n}{model}\PY{o}{.}\PY{n}{add}\PY{p}{(}\PY{n}{tf}\PY{o}{.}\PY{n}{keras}\PY{o}{.}\PY{n}{layers}\PY{o}{.}\PY{n}{Flatten}\PY{p}{(}\PY{p}{)}\PY{p}{)}
    \PY{n}{model}\PY{o}{.}\PY{n}{add}\PY{p}{(}\PY{n}{tf}\PY{o}{.}\PY{n}{keras}\PY{o}{.}\PY{n}{layers}\PY{o}{.}\PY{n}{Dense}\PY{p}{(}\PY{l+m+mi}{128}\PY{p}{,} \PY{n}{activation}\PY{o}{=}\PY{l+s+s1}{\PYZsq{}}\PY{l+s+s1}{relu}\PY{l+s+s1}{\PYZsq{}}\PY{p}{)}\PY{p}{)}
    \PY{n}{model}\PY{o}{.}\PY{n}{add}\PY{p}{(}\PY{n}{tf}\PY{o}{.}\PY{n}{keras}\PY{o}{.}\PY{n}{layers}\PY{o}{.}\PY{n}{Dropout}\PY{p}{(}\PY{l+m+mf}{0.1}\PY{p}{)}\PY{p}{)}
    \PY{n}{model}\PY{o}{.}\PY{n}{add}\PY{p}{(}\PY{n}{tf}\PY{o}{.}\PY{n}{keras}\PY{o}{.}\PY{n}{layers}\PY{o}{.}\PY{n}{Dense}\PY{p}{(}\PY{l+m+mi}{64}\PY{p}{,} \PY{n}{activation}\PY{o}{=}\PY{l+s+s1}{\PYZsq{}}\PY{l+s+s1}{relu}\PY{l+s+s1}{\PYZsq{}}\PY{p}{)}\PY{p}{)}
    \PY{n}{model}\PY{o}{.}\PY{n}{add}\PY{p}{(}\PY{n}{tf}\PY{o}{.}\PY{n}{keras}\PY{o}{.}\PY{n}{layers}\PY{o}{.}\PY{n}{Dense}\PY{p}{(}\PY{l+m+mi}{10}\PY{p}{,} \PY{n}{activation}\PY{o}{=}\PY{l+s+s1}{\PYZsq{}}\PY{l+s+s1}{softmax}\PY{l+s+s1}{\PYZsq{}}\PY{p}{)}\PY{p}{)}

    \PY{n}{model}\PY{o}{.}\PY{n}{compile}\PY{p}{(}\PY{n}{optimizer}\PY{o}{=}\PY{l+s+s2}{\PYZdq{}}\PY{l+s+s2}{rmsprop}\PY{l+s+s2}{\PYZdq{}}\PY{p}{,} \PY{n}{loss}\PY{o}{=}\PY{l+s+s2}{\PYZdq{}}\PY{l+s+s2}{categorical\PYZus{}crossentropy}\PY{l+s+s2}{\PYZdq{}}\PY{p}{,} \PY{n}{metrics}\PY{o}{=}\PY{p}{[}\PY{l+s+s2}{\PYZdq{}}\PY{l+s+s2}{accuracy}\PY{l+s+s2}{\PYZdq{}}\PY{p}{]}\PY{p}{)}
    
    \PY{k}{return} \PY{n}{shfl}\PY{o}{.}\PY{n}{model}\PY{o}{.}\PY{n}{DeepLearningModel}\PY{p}{(}\PY{n}{model}\PY{p}{)}
\end{Verbatim}
\end{tcolorbox}

The following step is the definition of the federated aggregation operator in order to complete the \textit{model selection} in FL. The framework provides some aggregation operators that we can use immediately and the possibility to define your own operator. In this case, we use the provided FedAvg operator.

    \begin{tcolorbox}[breakable, size=fbox, boxrule=1pt, pad at break*=1mm,colback=cellbackground, colframe=cellborder]
\prompt{In}{incolor}{7}{\boxspacing}
\begin{Verbatim}[commandchars=\\\{\}]
\PY{n}{aggregator} \PY{o}{=} \PY{n}{shfl}\PY{o}{.}\PY{n}{federated\PYZus{}aggregator}\PY{o}{.}\PY{n}{FedAvgAggregator}\PY{p}{(}\PY{p}{)}
\PY{n}{federated\PYZus{}government} \PY{o}{=} \PY{n}{shfl}\PY{o}{.}\PY{n}{federated\PYZus{}government}\PY{o}{.}\PY{n}{FederatedGovernment}\PY{p}{(}\PY{n}{model\PYZus{}builder}\PY{p}{,} \PY{n}{federated\PYZus{}data}\PY{p}{,} \PY{n}{aggregator}\PY{p}{)}
\end{Verbatim}
\end{tcolorbox}

The framework also provides the possibility of making data transformation for the \textit{data preprocessing} step, defining federated operations using \texttt{FederatedTransformation} interface. We first reshape data and then normalise it using test data mean and standard deviation (std) as normalisation parameters.

    \begin{tcolorbox}[breakable, size=fbox, boxrule=1pt, pad at break*=1mm,colback=cellbackground, colframe=cellborder]
\prompt{In}{incolor}{8}{\boxspacing}
\begin{Verbatim}[commandchars=\\\{\}]
\PY{k+kn}{import} \PY{n+nn}{numpy} \PY{k}{as} \PY{n+nn}{np}

\PY{k}{class} \PY{n+nc}{Reshape}\PY{p}{(}\PY{n}{shfl}\PY{o}{.}\PY{n}{private}\PY{o}{.}\PY{n}{FederatedTransformation}\PY{p}{)}\PY{p}{:}
    
    \PY{k}{def} \PY{n+nf}{apply}\PY{p}{(}\PY{n+nb+bp}{self}\PY{p}{,} \PY{n}{labeled\PYZus{}data}\PY{p}{)}\PY{p}{:}
        \PY{n}{labeled\PYZus{}data}\PY{o}{.}\PY{n}{data} \PY{o}{=} \PY{n}{np}\PY{o}{.}\PY{n}{reshape}\PY{p}{(}\PY{n}{labeled\PYZus{}data}\PY{o}{.}\PY{n}{data}\PY{p}{,} \PY{p}{(}\PY{n}{labeled\PYZus{}data}\PY{o}{.}\PY{n}{data}\PY{o}{.}\PY{n}{shape}\PY{p}{[}\PY{l+m+mi}{0}\PY{p}{]}\PY{p}{,} \PY{n}{labeled\PYZus{}data}\PY{o}{.}\PY{n}{data}\PY{o}{.}\PY{n}{shape}\PY{p}{[}\PY{l+m+mi}{1}\PY{p}{]}\PY{p}{,} \PY{n}{labeled\PYZus{}data}\PY{o}{.}\PY{n}{data}\PY{o}{.}\PY{n}{shape}\PY{p}{[}\PY{l+m+mi}{2}\PY{p}{]}\PY{p}{,}\PY{l+m+mi}{1}\PY{p}{)}\PY{p}{)}
        
\PY{n}{shfl}\PY{o}{.}\PY{n}{private}\PY{o}{.}\PY{n}{federated\PYZus{}operation}\PY{o}{.}\PY{n}{apply\PYZus{}federated\PYZus{}transformation}\PY{p}{(}\PY{n}{federated\PYZus{}data}\PY{p}{,} \PY{n}{Reshape}\PY{p}{(}\PY{p}{)}\PY{p}{)}
\end{Verbatim}
\end{tcolorbox}

    \begin{tcolorbox}[breakable, size=fbox, boxrule=1pt, pad at break*=1mm,colback=cellbackground, colframe=cellborder]
\prompt{In}{incolor}{9}{\boxspacing}
\begin{Verbatim}[commandchars=\\\{\}]
\PY{k+kn}{import} \PY{n+nn}{numpy} \PY{k}{as} \PY{n+nn}{np}

\PY{k}{class} \PY{n+nc}{Normalize}\PY{p}{(}\PY{n}{shfl}\PY{o}{.}\PY{n}{private}\PY{o}{.}\PY{n}{FederatedTransformation}\PY{p}{)}\PY{p}{:}
    
    \PY{k}{def} \PY{n+nf+fm}{\PYZus{}\PYZus{}init\PYZus{}\PYZus{}}\PY{p}{(}\PY{n+nb+bp}{self}\PY{p}{,} \PY{n}{mean}\PY{p}{,} \PY{n}{std}\PY{p}{)}\PY{p}{:}
        \PY{n+nb+bp}{self}\PY{o}{.}\PY{n}{\PYZus{}\PYZus{}mean} \PY{o}{=} \PY{n}{mean}
        \PY{n+nb+bp}{self}\PY{o}{.}\PY{n}{\PYZus{}\PYZus{}std} \PY{o}{=} \PY{n}{std}
    
    \PY{k}{def} \PY{n+nf}{apply}\PY{p}{(}\PY{n+nb+bp}{self}\PY{p}{,} \PY{n}{labeled\PYZus{}data}\PY{p}{)}\PY{p}{:}
        \PY{n}{labeled\PYZus{}data}\PY{o}{.}\PY{n}{data} \PY{o}{=} \PY{p}{(}\PY{n}{labeled\PYZus{}data}\PY{o}{.}\PY{n}{data} \PY{o}{\PYZhy{}} \PY{n+nb+bp}{self}\PY{o}{.}\PY{n}{\PYZus{}\PYZus{}mean}\PY{p}{)}\PY{o}{/}\PY{n+nb+bp}{self}\PY{o}{.}\PY{n}{\PYZus{}\PYZus{}std}
        
        
\PY{n}{mean} \PY{o}{=} \PY{n}{np}\PY{o}{.}\PY{n}{mean}\PY{p}{(}\PY{n}{train\PYZus{}data}\PY{o}{.}\PY{n}{data}\PY{p}{)}
\PY{n}{std} \PY{o}{=} \PY{n}{np}\PY{o}{.}\PY{n}{std}\PY{p}{(}\PY{n}{train\PYZus{}data}\PY{o}{.}\PY{n}{data}\PY{p}{)}
\PY{n}{shfl}\PY{o}{.}\PY{n}{private}\PY{o}{.}\PY{n}{federated\PYZus{}operation}\PY{o}{.}\PY{n}{apply\PYZus{}federated\PYZus{}transformation}\PY{p}{(}\PY{n}{federated\PYZus{}data}\PY{p}{,} \PY{n}{Normalize}\PY{p}{(}\PY{n}{mean}\PY{p}{,} \PY{n}{std}\PY{p}{)}\PY{p}{)}
\end{Verbatim}
\end{tcolorbox}

We are now ready to train the \paco{FL} algorithm. We run 2 rounds of learning showing test accuracy and loss of each client and test accuracy and loss of the global aggregated model.

    \begin{tcolorbox}[breakable, size=fbox, boxrule=1pt, pad at break*=1mm,colback=cellbackground, colframe=cellborder]
\prompt{In}{incolor}{10}{\boxspacing}
\begin{Verbatim}[commandchars=\\\{\}]
\PY{n}{test\PYZus{}data} \PY{o}{=} \PY{n}{np}\PY{o}{.}\PY{n}{reshape}\PY{p}{(}\PY{n}{test\PYZus{}data}\PY{p}{,} \PY{p}{(}\PY{n}{test\PYZus{}data}\PY{o}{.}\PY{n}{shape}\PY{p}{[}\PY{l+m+mi}{0}\PY{p}{]}\PY{p}{,} \PY{n}{test\PYZus{}data}\PY{o}{.}\PY{n}{shape}\PY{p}{[}\PY{l+m+mi}{1}\PY{p}{]}\PY{p}{,} \PY{n}{test\PYZus{}data}\PY{o}{.}\PY{n}{shape}\PY{p}{[}\PY{l+m+mi}{2}\PY{p}{]}\PY{p}{,}\PY{l+m+mi}{1}\PY{p}{)}\PY{p}{)}
\PY{n}{federated\PYZus{}government}\PY{o}{.}\PY{n}{run\PYZus{}rounds}\PY{p}{(}\PY{l+m+mi}{2}\PY{p}{,} \PY{n}{test\PYZus{}data}\PY{p}{,} \PY{n}{test\PYZus{}labels}\PY{p}{)}
\end{Verbatim}
\end{tcolorbox}

\begin{Verbatim}[commandchars=\\\{\}]
Accuracy round 0
Test performance client <shfl.private.federated\_operation.FederatedDataNode
object at 0x1a485e2450>: \textbf{[15.087034225463867, 0.9314000010490417]}
Test performance client <shfl.private.federated\_operation.FederatedDataNode
object at 0x106a0ffd0>: \textbf{[21.040000915527344, 0.9094250202178955]}
Test performance client <shfl.private.federated\_operation.FederatedDataNode
object at 0x1a48639e90>: \textbf{[11.712089538574219, 0.9425749778747559]}
Test performance client <shfl.private.federated\_operation.FederatedDataNode
object at 0x1a486396d0>: \textbf{[10.11756420135498, 0.9498249888420105]}
Test performance client <shfl.private.federated\_operation.FederatedDataNode
object at 0x1a48639c50>: \textbf{[24.04242706298828, 0.8968499898910522]}

\end{Verbatim}

\begin{tcolorbox}[breakable, size=fbox, boxrule=1pt, pad at break*=1mm,colback=white, colframe=cellborder]
Global model test performance :\textbf{ [7.954472064971924, 0.9403749704360962]}
\end{tcolorbox}

\begin{Verbatim}[commandchars=\\\{\}]
Accuracy round 1
Test performance client <shfl.private.federated\_operation.FederatedDataNode
object at 0x1a485e2450>: \textbf{[21.94520378112793, 0.9227499961853027]}
Test performance client <shfl.private.federated\_operation.FederatedDataNode
object at 0x106a0ffd0>: \textbf{[16.780630111694336, 0.9445000290870667]}
Test performance client <shfl.private.federated\_operation.FederatedDataNode
object at 0x1a48639e90>: \textbf{[13.413337707519531, 0.9463250041007996]}
Test performance client <shfl.private.federated\_operation.FederatedDataNode
object at 0x1a486396d0>: \textbf{[9.085938453674316, 0.9628000259399414]}
Test performance client <shfl.private.federated\_operation.FederatedDataNode
object at 0x1a48639c50>: \textbf{[20.926694869995117, 0.918524980545044]}


\end{Verbatim}

\begin{tcolorbox}[breakable, size=fbox, boxrule=1pt, pad at break*=1mm,colback=white, colframe=cellborder]
Global model test performance : \textbf{[10.171743392944336, 0.958299994468689]}
\end{tcolorbox}

If we focus our attention on test accuracy in each client, we realise that there are widely varying results.\footnote{Running more learning rounds results in better performance as in the next section. The purpose of this example is to show how it works.} This is because of the scattered nature of the data distribution, which causes disparity in the quality of training data between clients.

\subsubsection{Comparison with a centralised convolutional neural network approach}
We analyse the behaviour of the FL approach in comparison with the equivalent centralised approach, which means training the neural network represented in Figure \ref{fig:cnn} on the same data using centralised learning. 


For this experiment, we use 25 clients and 10 rounds of learning with 5 epochs in both IID and non-IID scenario, where the nodes' data contain only a portion of all labels. For a fair comparison, in the classical approach we train for $epochs_{FL} \times rounds_{FL}$ epochs.

\begin{table}[hbt]
\centering
\begin{tabular}{@{}lrr@{}} \toprule
                            & \textbf{IID}      & \textbf{non-IID} \\ \midrule
\textbf{Centralised approach} & 0.9904           & \textbf{0.9901} \\
\textbf{Federated approach} & \textbf{0.9921}   & 0.9855          \\  \bottomrule
\end{tabular}
\caption{Accuracy of the FL and the classical approach, in both IID and non-IID scenarios. 
In the FL case, the data is distributed over 25 clients, and 10 FL rounds of learning with 5 epochs per client are employed.
}\label{tab:tabla} 
\end{table}


The high performance of the federated approach stands out in Table \ref{tab:tabla}, where the accuracy for the considered scenarios is reported. In the IID scenario, it beats the centralised approach results, which shows the robustness of the approach caused by the combination of the information learned by each client. In the non-IID scenario, the federated approach attains lower results than the centralised one due to the additional challenge of non-homogeneous distribution of data across clients.\footnote{The performance of the centralised approach using non-IID data is not perfectly identical to the IID case due to the random sampling employed when generating the non-IID nodes' data.} However, the results are very competitive highlighting the strength of the federated approach.

\subsection{Linear regression with DP}\label{sec:linear_regressionDP}

This section presents a linear regression FL simulation with DP \correcciones{following the methodological guidelines with} \sherpa{\texttt{Sherpa.ai} FL}.\footnote{\sherpa{\texttt{Sherpa.ai} FL} offers support for the linear regression model from  \textrm{scikit-learn} \url{https://scikit-learn.org/stable/index.html}} 
The Laplace mechanism is used when the model's sensitivity is estimated by a sampling procedure \citep{Rubinstein2017}.
Moreover, we demonstrate the application of the advanced composition theorem for DP  for not exceeding the maximum privacy loss allowed (see Section \ref{sec:dp_key_elements}). 

\subsubsection{Case of study}
We will use the California Housing dataset,\footnote{\url{https://scikit-learn.org/stable/modules/generated/sklearn.datasets.fetch_california_housing.html}} which consists of approximately 20\,000 samples for median house prices in California.
Although the dataset possesses eight features, in this example we will only make use of the first two, in order to reduce the variance in the prediction.
The (single) target is the cost of the house.
As it can be observed in the code below, we retain 2\,000 samples for later use with the sensitivity sampling for DP, and the rest of the data is split in train and test sets 
as detailed in Table \ref{tab:lr_train_test}.

\begin{table}[hbt]
\centering
\begin{tabular}{@{}rrr@{}} \toprule
\textbf{Train set} & \textbf{Test set} & \textbf{Total} \\\midrule
14\,912 & 3\,728 & 18\,640 \\
 \bottomrule
\end{tabular}
\caption{Distribution of the California Housing dataset.}
\label{tab:lr_train_test} 
\end{table}

For the FL simulation  we use 5 clients among which the train dataset is IID. FedAvg is chosen as the federated aggregation operator. The code of the example is detailed in the following section.

    \hypertarget{description-of-the-code}{%
\subsubsection{Description of the code}\label{description-of-the-code}}

\texttt{Sherpa.FL} allows to easily convert a generic dataset to
interact with the platform:

    \begin{tcolorbox}[breakable, size=fbox, boxrule=1pt, pad at break*=1mm,colback=cellbackground, colframe=cellborder]
\begin{Verbatim}[commandchars=\\\{\}]
\PY{k+kn}{import} \PY{n+nn}{shfl}
\PY{k+kn}{from} \PY{n+nn}{shfl}\PY{n+nn}{.}\PY{n+nn}{data\PYZus{}base}\PY{n+nn}{.}\PY{n+nn}{data\PYZus{}base} \PY{k+kn}{import} \PY{n}{LabeledDatabase}
\PY{k+kn}{import} \PY{n+nn}{sklearn}\PY{n+nn}{.}\PY{n+nn}{datasets}
\PY{k+kn}{import} \PY{n+nn}{numpy} \PY{k}{as} \PY{n+nn}{np}
\PY{k+kn}{from} \PY{n+nn}{shfl}\PY{n+nn}{.}\PY{n+nn}{private}\PY{n+nn}{.}\PY{n+nn}{reproducibility} \PY{k+kn}{import} \PY{n}{Reproducibility}

\PY{c+c1}{\PYZsh{} Comment to turn off reproducibility:}
\PY{n}{Reproducibility}\PY{p}{(}\PY{l+m+mi}{1234}\PY{p}{)}

\PY{n}{all\PYZus{}data} \PY{o}{=} \PY{n}{sklearn}\PY{o}{.}\PY{n}{datasets}\PY{o}{.}\PY{n}{fetch\PYZus{}california\PYZus{}housing}\PY{p}{(}\PY{p}{)}
\PY{n}{n\PYZus{}features} \PY{o}{=} \PY{l+m+mi}{2}
\PY{n}{data} \PY{o}{=} \PY{n}{all\PYZus{}data}\PY{p}{[}\PY{l+s+s2}{\PYZdq{}}\PY{l+s+s2}{data}\PY{l+s+s2}{\PYZdq{}}\PY{p}{]}\PY{p}{[}\PY{p}{:}\PY{p}{,}\PY{l+m+mi}{0}\PY{p}{:}\PY{n}{n\PYZus{}features}\PY{p}{]}
\PY{n}{labels} \PY{o}{=} \PY{n}{all\PYZus{}data}\PY{p}{[}\PY{l+s+s2}{\PYZdq{}}\PY{l+s+s2}{target}\PY{l+s+s2}{\PYZdq{}}\PY{p}{]}    

\PY{c+c1}{\PYZsh{} Retain part for DP sensitivity sampling:}
\PY{n}{size} \PY{o}{=} \PY{l+m+mi}{2000}
\PY{n}{sampling\PYZus{}data} \PY{o}{=} \PY{n}{data}\PY{p}{[}\PY{o}{\PYZhy{}}\PY{n}{size}\PY{p}{:}\PY{p}{,} \PY{p}{]}
\PY{n}{sampling\PYZus{}labels} \PY{o}{=} \PY{n}{labels}\PY{p}{[}\PY{o}{\PYZhy{}}\PY{n}{size}\PY{p}{:}\PY{p}{,} \PY{p}{]}

\PY{c+c1}{\PYZsh{} Create database:}
\PY{n}{database} \PY{o}{=} \PY{n}{LabeledDatabase}\PY{p}{(}\PY{n}{data}\PY{p}{[}\PY{l+m+mi}{0}\PY{p}{:}\PY{o}{\PYZhy{}}\PY{n}{size}\PY{p}{,} \PY{p}{]}\PY{p}{,} \PY{n}{labels}\PY{p}{[}\PY{l+m+mi}{0}\PY{p}{:}\PY{o}{\PYZhy{}}\PY{n}{size}\PY{p}{]}\PY{p}{)}
\PY{n}{train\PYZus{}data}\PY{p}{,} \PY{n}{train\PYZus{}labels}\PY{p}{,} \PY{n}{test\PYZus{}data}\PY{p}{,} \PY{n}{test\PYZus{}labels} \PY{o}{=} \PY{n}{database}\PY{o}{.}\PY{n}{load\PYZus{}data}\PY{p}{(}\PY{p}{)}
\end{Verbatim}
\end{tcolorbox}

    We will simulate a FL scenario by distributing the train data over a
collection of clients, assuming an IID setting:

    \begin{tcolorbox}[breakable, size=fbox, boxrule=1pt, pad at break*=1mm,colback=cellbackground, colframe=cellborder]
\begin{Verbatim}[commandchars=\\\{\}]
\PY{n}{iid\PYZus{}distribution} \PY{o}{=} \PY{n}{shfl}\PY{o}{.}\PY{n}{data\PYZus{}distribution}\PY{o}{.}\PY{n}{IidDataDistribution}\PY{p}{(}\PY{n}{database}\PY{p}{)}
\PY{n}{federated\PYZus{}data}\PY{p}{,} \PY{n}{test\PYZus{}data}\PY{p}{,} \PY{n}{test\PYZus{}labels} \PY{o}{=} \PY{n}{iid\PYZus{}distribution}\PY{o}{.}\PY{n}{get\PYZus{}federated\PYZus{}data}\PY{p}{(}\PY{n}{num\PYZus{}nodes}\PY{o}{=}\PY{l+m+mi}{5}\PY{p}{)}
\end{Verbatim}
\end{tcolorbox}

    At this stage, we need to define the linear regression model, and we
choose the \correcciones{aggregation operator} to be the average of the clients'
models:

    \begin{tcolorbox}[breakable, size=fbox, boxrule=1pt, pad at break*=1mm,colback=cellbackground, colframe=cellborder]
\begin{Verbatim}[commandchars=\\\{\}]
\PY{k+kn}{from} \PY{n+nn}{shfl}\PY{n+nn}{.}\PY{n+nn}{model}\PY{n+nn}{.}\PY{n+nn}{linear\PYZus{}regression\PYZus{}model} \PY{k+kn}{import} \PY{n}{LinearRegressionModel}

\PY{k}{def} \PY{n+nf}{model\PYZus{}builder}\PY{p}{(}\PY{p}{)}\PY{p}{:}
    \PY{n}{model} \PY{o}{=} \PY{n}{LinearRegressionModel}\PY{p}{(}\PY{n}{n\PYZus{}features}\PY{o}{=}\PY{n}{n\PYZus{}features}\PY{p}{)}
    \PY{k}{return} \PY{n}{model}

\PY{n}{aggregator} \PY{o}{=} \PY{n}{shfl}\PY{o}{.}\PY{n}{federated\PYZus{}aggregator}\PY{o}{.}\PY{n}{FedAvgAggregator}\PY{p}{(}\PY{p}{)}
\end{Verbatim}
\end{tcolorbox}

    \hypertarget{running-the-model-in-a-federated-configuration}{%
\subsubsection{Running the model in a Federated
configuration}\label{running-the-model-in-a-federated-configuration}}

We are now ready to run the FL model. Note that in this case, we set the
number of rounds \texttt{n=1} since no iterations are needed in the case
of linear regression. The performance metrics used are the Root Mean
Squared Error (RMSE) and the \(R^2\) score. It can be observed that the
performance of the \emph{Global model} (i.e.~the aggregated model) is in
general superior with respect to the performance of each node, thus the
federated learning approach proves to be beneficial:

    \begin{tcolorbox}[breakable, size=fbox, boxrule=1pt, pad at break*=1mm,colback=cellbackground, colframe=cellborder]
\begin{Verbatim}[commandchars=\\\{\}]
\PY{n}{federated\PYZus{}government} \PY{o}{=} \PY{n}{shfl}\PY{o}{.}\PY{n}{federated\PYZus{}government}\PY{o}{.}\PY{n}{FederatedGovernment}\PY{p}{(}\PY{n}{model\PYZus{}builder}\PY{p}{,} \PY{n}{federated\PYZus{}data}\PY{p}{,} \PY{n}{aggregator}\PY{p}{)}
\PY{n}{federated\PYZus{}government}\PY{o}{.}\PY{n}{run\PYZus{}rounds}\PY{p}{(}\PY{n}{n}\PY{o}{=}\PY{l+m+mi}{1}\PY{p}{,} \PY{n}{test\PYZus{}data}\PY{o}{=}\PY{n}{test\PYZus{}data}\PY{p}{,} \PY{n}{test\PYZus{}label}\PY{o}{=}\PY{n}{test\PYZus{}labels}\PY{p}{)}
\end{Verbatim}
\end{tcolorbox}

    \begin{Verbatim}[commandchars=\\\{\}]
Accuracy round 0
Test performance client <shfl.private.federated\_operation.FederatedDataNode object at 0x7f63606a08d0>:
\textbf{[0.8161535463006577, 0.5010049851923566]}
Test performance client <shfl.private.federated\_operation.FederatedDataNode object at 0x7f63606a0ac8>:
\textbf{[0.81637303674763, 0.5007365568636023]}
Test performance client <shfl.private.federated\_operation.FederatedDataNode object at 0x7f63606a09b0>:
\textbf{[0.8155342443231007, 0.5017619784187599]}
Test performance client <shfl.private.federated\_operation.FederatedDataNode object at 0x7f63606a0be0>:
\textbf{[0.8158502097728687, 0.5013758352304256]}
Test performance client <shfl.private.federated\_operation.FederatedDataNode object at 0x7f63606a0cf8>:
\textbf{[0.8151607067608612, 0.5022182878756591]}
    \end{Verbatim}
\begin{tcolorbox}[breakable, size=fbox, boxrule=1pt, pad at break*=1mm,colback=white, colframe=cellborder]
Global model test performance : \textbf{[0.8154147770321544, 0.5019079411109164]}
\end{tcolorbox}

    \hypertarget{differential-privacy-sampling-the-models-sensitivity}{%
\subsubsection{Differential Privacy: sampling the model's
sensitivity}\label{differential-privacy-sampling-the-models-sensitivity}}

In the case of applying the Laplace privacy mechanism (see Section
\ref{sec:dp_key_elements}), the noise added has to be of the order
of the sensitivity of the model's output, i.e.~the model parameters of
our linear regression. In the general case, the model's sensitivity
might be difficult to compute analytically. An alternative approach is
to attain \emph{random} differential privacy through a sampling over the
data \citep{Rubinstein2017}. That is, instead of computing analytically
the \emph{global} sensitivity \(\Delta f\), we compute an
\emph{empirical estimation} of it by sampling over the dataset. This
approach is convenient since it allows for the sensitivity estimation of
an arbitrary model or a black-box computer function. The
\texttt{Sherpa.FL} framework provides this functionality in the class
\texttt{SensitivitySampler}.

In order to carry out this approach, we need to specify a distribution
of the data to sample from. This in general requires previous knowledge
and/or model assumptions. In order not make any specific assumption on
the distribution of the dataset, we can choose a \emph{uniform}
distribution. To the end, we define our class of
\texttt{ProbabilityDistribution} that uniformly samples over a
data-frame. We use the previously retained part of the dataset for
sampling:

    \begin{tcolorbox}[breakable, size=fbox, boxrule=1pt, pad at break*=1mm,colback=cellbackground, colframe=cellborder]
\begin{Verbatim}[commandchars=\\\{\}]
\PY{k}{class} \PY{n+nc}{UniformDistribution}\PY{p}{(}\PY{n}{shfl}\PY{o}{.}\PY{n}{differential\PYZus{}privacy}\PY{o}{.}\PY{n}{ProbabilityDistribution}\PY{p}{)}\PY{p}{:}
    \PY{l+s+sd}{\PYZdq{}\PYZdq{}\PYZdq{}}
\PY{l+s+sd}{    Implement Uniform sampling over the data}
\PY{l+s+sd}{    \PYZdq{}\PYZdq{}\PYZdq{}}
    \PY{k}{def} \PY{n+nf+fm}{\PYZus{}\PYZus{}init\PYZus{}\PYZus{}}\PY{p}{(}\PY{n+nb+bp}{self}\PY{p}{,} \PY{n}{sample\PYZus{}data}\PY{p}{)}\PY{p}{:}
        \PY{n+nb+bp}{self}\PY{o}{.}\PY{n}{\PYZus{}sample\PYZus{}data} \PY{o}{=} \PY{n}{sample\PYZus{}data}

    \PY{k}{def} \PY{n+nf}{sample}\PY{p}{(}\PY{n+nb+bp}{self}\PY{p}{,} \PY{n}{sample\PYZus{}size}\PY{p}{)}\PY{p}{:}
        \PY{n}{row\PYZus{}indices} \PY{o}{=} \PY{n}{np}\PY{o}{.}\PY{n}{random}\PY{o}{.}\PY{n}{randint}\PY{p}{(}\PY{n}{low}\PY{o}{=}\PY{l+m+mi}{0}\PY{p}{,} \PY{n}{high}\PY{o}{=}\PY{n+nb+bp}{self}\PY{o}{.}\PY{n}{\PYZus{}sample\PYZus{}data}\PY{o}{.}\PY{n}{shape}\PY{p}{[}\PY{l+m+mi}{0}\PY{p}{]}\PY{p}{,} \PY{n}{size}\PY{o}{=}\PY{n}{sample\PYZus{}size}\PY{p}{,} \PY{n}{dtype}\PY{o}{=}\PY{l+s+s1}{\PYZsq{}}\PY{l+s+s1}{l}\PY{l+s+s1}{\PYZsq{}}\PY{p}{)}
        
        \PY{k}{return} \PY{n+nb+bp}{self}\PY{o}{.}\PY{n}{\PYZus{}sample\PYZus{}data}\PY{p}{[}\PY{n}{row\PYZus{}indices}\PY{p}{,} \PY{p}{:}\PY{p}{]}
    
\PY{n}{sample\PYZus{}data} \PY{o}{=} \PY{n}{np}\PY{o}{.}\PY{n}{hstack}\PY{p}{(}\PY{p}{(}\PY{n}{sampling\PYZus{}data}\PY{p}{,} \PY{n}{sampling\PYZus{}labels}\PY{o}{.}\PY{n}{reshape}\PY{p}{(}\PY{o}{\PYZhy{}}\PY{l+m+mi}{1}\PY{p}{,}\PY{l+m+mi}{1}\PY{p}{)}\PY{p}{)}\PY{p}{)}
\end{Verbatim}
\end{tcolorbox}

    The class \texttt{SensitivitySampler} implements the sampling given a
\emph{query}, i.e.~the learning model itself in this case. We only need
to add the method \texttt{get} to our model since it is required by the
class \texttt{SensitivitySampler}. We choose the sensitivity norm to be
the $\ell_1$ norm and we apply the sampling. The value of the sensitivity
depends on the number of samples \texttt{n}: the more samples we
perform, the more accurate the sensitivity. Indeed, increasing the
number of samples \texttt{n}, the sensitivity gets more accurate and
typically decreases.

    \begin{tcolorbox}[breakable, size=fbox, boxrule=1pt, pad at break*=1mm,colback=cellbackground, colframe=cellborder]
\begin{Verbatim}[commandchars=\\\{\}]
\PY{k+kn}{from} \PY{n+nn}{shfl}\PY{n+nn}{.}\PY{n+nn}{differential\PYZus{}privacy} \PY{k+kn}{import} \PY{n}{SensitivitySampler}
\PY{k+kn}{from} \PY{n+nn}{shfl}\PY{n+nn}{.}\PY{n+nn}{differential\PYZus{}privacy} \PY{k+kn}{import} \PY{n}{L1SensitivityNorm}

\PY{k}{class} \PY{n+nc}{LinearRegressionSample}\PY{p}{(}\PY{n}{LinearRegressionModel}\PY{p}{)}\PY{p}{:}
    
    \PY{k}{def} \PY{n+nf}{get}\PY{p}{(}\PY{n+nb+bp}{self}\PY{p}{,} \PY{n}{data\PYZus{}array}\PY{p}{)}\PY{p}{:}
        \PY{n}{data} \PY{o}{=} \PY{n}{data\PYZus{}array}\PY{p}{[}\PY{p}{:}\PY{p}{,} \PY{l+m+mi}{0}\PY{p}{:}\PY{o}{\PYZhy{}}\PY{l+m+mi}{1}\PY{p}{]}
        \PY{n}{labels} \PY{o}{=} \PY{n}{data\PYZus{}array}\PY{p}{[}\PY{p}{:}\PY{p}{,} \PY{o}{\PYZhy{}}\PY{l+m+mi}{1}\PY{p}{]}
        \PY{n}{train\PYZus{}model} \PY{o}{=} \PY{n+nb+bp}{self}\PY{o}{.}\PY{n}{train}\PY{p}{(}\PY{n}{data}\PY{p}{,} \PY{n}{labels}\PY{p}{)}
      
        \PY{k}{return} \PY{n+nb+bp}{self}\PY{o}{.}\PY{n}{get\PYZus{}model\PYZus{}params}\PY{p}{(}\PY{p}{)}

\PY{n}{distribution} \PY{o}{=} \PY{n}{UniformDistribution}\PY{p}{(}\PY{n}{sample\PYZus{}data}\PY{p}{)}
\PY{n}{sampler} \PY{o}{=} \PY{n}{SensitivitySampler}\PY{p}{(}\PY{p}{)}
\PY{n}{n\PYZus{}samples} \PY{o}{=} \PY{l+m+mi}{4000}
\PY{n}{max\PYZus{}sensitivity}\PY{p}{,} \PY{n}{mean\PYZus{}sensitivity} \PY{o}{=} \PY{n}{sampler}\PY{o}{.}\PY{n}{sample\PYZus{}sensitivity}\PY{p}{(}
    \PY{n}{LinearRegressionSample}\PY{p}{(}\PY{n}{n\PYZus{}features}\PY{o}{=}\PY{n}{n\PYZus{}features}\PY{p}{,} \PY{n}{n\PYZus{}targets}\PY{o}{=}\PY{l+m+mi}{1}\PY{p}{)}\PY{p}{,} 
    \PY{n}{L1SensitivityNorm}\PY{p}{(}\PY{p}{)}\PY{p}{,} \PY{n}{distribution}\PY{p}{,} \PY{n}{n}\PY{o}{=}\PY{n}{n\PYZus{}samples}\PY{p}{,} \PY{n}{gamma}\PY{o}{=}\PY{l+m+mf}{0.05}\PY{p}{)}
\PY{n+nb}{print}\PY{p}{(}\PY{l+s+s2}{\PYZdq{}}\PY{l+s+s2}{Max sensitivity from sampling: }\PY{l+s+s2}{\PYZdq{}} \PY{o}{+} \PY{n+nb}{str}\PY{p}{(}\PY{n}{max\PYZus{}sensitivity}\PY{p}{)}\PY{p}{)}
\PY{n+nb}{print}\PY{p}{(}\PY{l+s+s2}{\PYZdq{}}\PY{l+s+s2}{Mean sensitivity from sampling: }\PY{l+s+s2}{\PYZdq{}} \PY{o}{+} \PY{n+nb}{str}\PY{p}{(}\PY{n}{mean\PYZus{}sensitivity}\PY{p}{)}\PY{p}{)}
\end{Verbatim}
\end{tcolorbox}

    \begin{Verbatim}[commandchars=\\\{\}]
Max sensitivity from sampling: 0.008294354064053988
Mean sensitivity from sampling: 0.0006633612087443363
    \end{Verbatim}

    Unfortunately, sampling over a dataset involves the
training of the model on two datasets differing in one entry
\citep{Rubinstein2017}. Thus in general this procedure might be
computationally expensive (e.g.~in the case of training a deep neuronal
network).

    \hypertarget{running-the-model-in-a-federated-configuration-with-differential-privacy}{%
\subsubsection{Running the model in a Federated configuration with
Differential
Privacy}\label{running-the-model-in-a-federated-configuration-with-differential-privacy}}

At this stage we are ready to add a layer of DP to our federated
learning model. Specifically, we will apply the Laplace mechanism from Section
\ref{sec:dp_key_elements}, employing the sensitivity obtained from
the previous sampling, namely \(\Delta f \approx 0.008\). The Laplace
mechanism provided by the \texttt{Sherpa.FL} framework is then assigned
as the \textit{private} access type to the model's parameters of each
client in a new \texttt{FederatedGovernment} object. This results into
an \(\epsilon\)-\textit{differentially private FL model}. For example,
picking the value \(\epsilon = 0.5\), we can run the FL experiment with
DP:

    \begin{tcolorbox}[breakable, size=fbox, boxrule=1pt, pad at break*=1mm,colback=cellbackground, colframe=cellborder]
\begin{Verbatim}[commandchars=\\\{\}]
\PY{k+kn}{from} \PY{n+nn}{shfl}\PY{n+nn}{.}\PY{n+nn}{differential\PYZus{}privacy} \PY{k+kn}{import} \PY{n}{LaplaceMechanism}

\PY{n}{params\PYZus{}access\PYZus{}definition} \PY{o}{=} \PY{n}{LaplaceMechanism}\PY{p}{(}\PY{n}{sensitivity}\PY{o}{=}\PY{n}{max\PYZus{}sensitivity}\PY{p}{,} \PY{n}{epsilon}\PY{o}{=}\PY{l+m+mf}{0.5}\PY{p}{)}
\PY{n}{federated\PYZus{}governmentDP} \PY{o}{=} \PY{n}{shfl}\PY{o}{.}\PY{n}{federated\PYZus{}government}\PY{o}{.}\PY{n}{FederatedGovernment}\PY{p}{(}
    \PY{n}{model\PYZus{}builder}\PY{p}{,} \PY{n}{federated\PYZus{}data}\PY{p}{,} \PY{n}{aggregator}\PY{p}{,} \PY{n}{model\PYZus{}params\PYZus{}access}\PY{o}{=}\PY{n}{params\PYZus{}access\PYZus{}definition}\PY{p}{)}
\PY{n}{federated\PYZus{}governmentDP}\PY{o}{.}\PY{n}{run\PYZus{}rounds}\PY{p}{(}\PY{n}{n}\PY{o}{=}\PY{l+m+mi}{1}\PY{p}{,} \PY{n}{test\PYZus{}data}\PY{o}{=}\PY{n}{test\PYZus{}data}\PY{p}{,} \PY{n}{test\PYZus{}label}\PY{o}{=}\PY{n}{test\PYZus{}labels}\PY{p}{)}
\end{Verbatim}
\end{tcolorbox}

    \begin{Verbatim}[commandchars=\\\{\}]
Accuracy round 0
Test performance client <shfl.private.federated\_operation.FederatedDataNode object at 0x7f63606a08d0>:
\textbf{[0.8161535463006577, 0.5010049851923566]}
Test performance client <shfl.private.federated\_operation.FederatedDataNode object at 0x7f63606a0ac8>:
\textbf{[0.81637303674763, 0.5007365568636023]}
Test performance client <shfl.private.federated\_operation.FederatedDataNode object at 0x7f63606a09b0>:
\textbf{[0.8155342443231007, 0.5017619784187599]}
Test performance client <shfl.private.federated\_operation.FederatedDataNode object at 0x7f63606a0be0>:
\textbf{[0.8158502097728687, 0.5013758352304256]}
Test performance client <shfl.private.federated\_operation.FederatedDataNode object at 0x7f63606a0cf8>:
\textbf{[0.8151607067608612, 0.5022182878756591]}
    \end{Verbatim}
\begin{tcolorbox}[breakable, size=fbox, boxrule=1pt, pad at break*=1mm,colback=white, colframe=cellborder]
Global model test performance : \textbf{[0.8309024800913748, 0.48280707735516126]}
\end{tcolorbox}

    In the above example we observed that the performance of the model has
slightly deteriorated due to the addition of DP. In general, the privacy
increases at expenses of accuracy (i.e.~for smaller values of
\(\epsilon\)).


\subsubsection{Comparison with centralised and non private approaches}

It is of practical interest to assess the performance loss due to DP in the FL context. 
Table \ref{tab:linear_regression_federatedDP} reports the performance metrics for the centralised model, for the \textit{FL  non-private} model, and for the \textit{FL differentially-private} model.
In all the cases, the  models have learned on the train set, and the performance results have been computed over the test set.
The data is IID over 5 clients in the FL cases. 
For the federated DP cases, different values of $\epsilon= \{0.2,0.5,0.8\}$ are used and the total privacy expense is limited at $\epsilon_T = 4$. Thus, for each case, we took the average over the total runs before the budget is expended (see discussion about advanced composition theorems for privacy filters in Section \ref{sec:dp_key_elements}).  The sensitivity is fixed, employing the value obtained from sampling above.\footnote{Note that, when applying the composition theorems for privacy filters in the present example, we are assuming that the estimated sensitivity is a good enough approximation of the analytic sensitivity \citep{privacyfilters}.}

\begin{table}[hbt]

\centering
\begin{tabular}{@{}lrlr@{}} \toprule
\textbf{Approach}   & \textbf{RMSE}   & $\mathbf{R^2}$ \\ \midrule
Classical                         & \textbf{0.81540}&\textbf{0.50192} \\
Federated non-private             & 0.81541  & 0.50190 \\  
Federated  DP ($\epsilon = 0.2$, average of 20 runs)    & 1.05541 & 0.04224     \\ 
Federated  DP ($\epsilon = 0.5$, average of 8 runs)     & 0.84501  & 0.46457     \\  
Federated  DP ($\epsilon = 0.8$, average of 5 runs)     & 0.82171  & 0.49414     \\
\bottomrule
\end{tabular}
\caption{Federated linear regression: comparison between the classical centralised model, the non-private FL model, and the FL model with a DP layer using the Laplace mechanism. For the DP cases, the results are the average over the total runs allowed for the maximum privacy budget $\epsilon_T = 4$. Different values of $\epsilon= \{0.2,0.5,0.8\}$ are considered, and the sensitivity is fixed. The data is IID distributed over 5 clients. }
\label{tab:linear_regression_federatedDP} 
\end{table}

It can be observed that the centralised model and the non-private FL model exhibit \textit{comparable performance}, thus the accuracy is not degraded by applying  a FL approach.
The application of the Laplace mechanism guarantees $\epsilon$-DP, and the accuracy of the FL model can be leveraged by setting the value of $\epsilon$: for higher values, lower privacy is guaranteed, but the accuracy increases.   

\section{Concluding remarks}\label{sec:conclusion}
\label{sec:conclusions}

\correcciones{The characteristics of FL and DP make them good candidates to support AI services at the edges and to preserve data privacy. Hence, several software tools for FL and DP have been released. After a comparative analysis, we conclude that these software tools do not provide a unified support for FL and DP, and they do not follow any particular methodological guidelines that direct the developing of AI services to preserve data privacy.

Since FL is a machine learning paradigm, we have studied how to adapt the machine learning principles to the FL ones, and consequently we have also defined the workflow of an experimental setting of FL. The main result of that study is \sherpa{\texttt{Sherpa.ai} FL}, which is a new software framework with a unified support for FL and DP, that allows to follow the defined methodological guidelines for FL.

The combination of the methodological guidelines and \sherpa{\texttt{Sherpa.ai} FL} is shown by means of a classification and a regression use cases. Those illustrative examples also show that the centralised and federated setting of the same experiments achieve similar results, which means that the joint use FL and DP can support the development of AI services at the edges that preserve data privacy.

\sherpa{\texttt{Sherpa.ai} FL} is in
continuous development. Since FL and DP fields are constantly growing, we plan to extend the framework's functionalities by new federated aggregation operators, machine learning models, and data distributions. Moreover, new DP mechanisms such as RAPPOR \citep{bib:rappor} will be added, together with relaxations of DP such as Concentrated DP \citep{bib:dwork2016concentrated} or Rényi DP \citep{bib:rdp}.

}

\section*{Acknowledgments}
\label{sec:acknowledgment}

This research work is partially supported by the contract OTRI-4137 with SHERPA Europe S.L., the Spanish Government project TIN2017-89517-P. Nuria Rodr\'{i}guez Barroso and Eugenio Mart\'{i}nez C\'{a}mara were supported by the Spanish Government fellowship programmes Formación de Profesorado Universitario (FPU18/04475) and Juan de la Cierva Incorporaci\'{o}n (IJC2018-036092-I) respectively.


\bibliography{main}
\bibliographystyle{elsarticle-num-names} 

\end{document}